\documentclass{article}
\usepackage{graphicx} 
\usepackage{hyperref}
\usepackage[square, numbers]{natbib}
\usepackage{geometry}
\usepackage{amsmath}
\usepackage{amsthm}
\usepackage{booktabs}
\usepackage{authblk}

\usepackage{subcaption}

\usepackage{siunitx} 

\usepackage{titletoc} 

\usepackage{color}

\usepackage{xcolor} 
\definecolor{promptcolor}{RGB}{76, 135, 184}
\definecolor{boxfillcolor}{HTML}{f1f1f1}
\definecolor{boxframecolor}{HTML}{ffffff}
\definecolor{lightgray}{HTML}{F3F3F3}

\newcommand{\ourwork}{\textbf{Cardiverse}}
\usepackage{listings}
\lstdefinelanguage{markdown}{
  moredelim=[is][\bfseries]{**}{**},
  moredelim=[s][\color{black}\ttfamily]{`}{`},
  moredelim=[is][\color{black}\bfseries]{[}{]},
  comment=[l]{<!--},
  commentstyle=\color{gray}\itshape,
  string=[b]",
  stringstyle=\color{black},
  sensitive=true,
}

\usepackage[most,listings]{tcolorbox}

\definecolor{todocolor}{HTML}{A10000}

\title{Cardiverse: Harnessing LLMs for Novel Card Game Prototyping}

\author[1]{Danrui Li}
\author[1]{Sen Zhang}
\author[1]{Samuel S. Sohn}
\author[1]{Kaidong Hu}
\author[2]{Muhammad Usman}
\author[1,3]{Mubbasir Kapadia}
\affil[1]{\small Rutgers University, NJ, USA}
\affil[2]{\small Ontario Tech University, ON, Canada}
\affil[3]{\small Roblox, San Meteo, CA, USA}

\date{}

\begin{document}
\markboth{Danrui Li, Sen Zhang, Samuel S. Sohn et al.}{Cardiverse}
\maketitle

\begin{abstract}

The prototyping of computer games, particularly card games, requires extensive human effort in creative ideation and gameplay evaluation. Recent advances in Large Language Models (LLMs) offer opportunities to automate and streamline these processes. However, it remains challenging for LLMs to design novel game mechanics beyond existing databases, generate consistent gameplay environments, and develop scalable gameplay AI for large-scale evaluations.
This paper addresses these challenges by introducing a comprehensive automated card game prototyping framework. 
The approach highlights a graph-based indexing method for generating novel game variations, an LLM-driven system for consistent game code generation validated by gameplay records, and a gameplay AI constructing method that uses an ensemble of LLM-generated heuristic functions optimized through self-play. These contributions aim to accelerate card game prototyping, reduce human labor, and lower barriers to entry for game developers.
Code repo: \href{https://github.com/danruili/Cardiverse}{https://github.com/danruili/Cardiverse}

\end{abstract}

\section{Introduction}
\label{sec:intro}

Prototyping is a central part of computer game development, where game designers create a working model of an initial idea that allows for downstream evaluations on gameplay experience~\citep{fullerton2014game}. It does not focus on product-level quality but fast iterations on core game mechanics and their implementations.
Extensive research has been conducted to automate or assist these processes through computational methods~\cite{Gallotta_2024}. 
Recent advances in Large Language Models (LLMs) have further expanded the possibilities, enhancing tasks such as game mechanic design~\cite{charity2023conceptualgame}, programming~\cite{wu2024instructiondriven,chatdev}, and game AI development~\cite{yao_react_2023}. 
LLMs show the potential to integrate these subtasks into a comprehensive game prototyping pipeline, especially for card games that can be represented purely in text.
However, LLMs face the following significant challenges~\cite{Gallotta_2024}.

\begin{figure}[t]
  \centering
  \includegraphics[width=.5\textwidth]{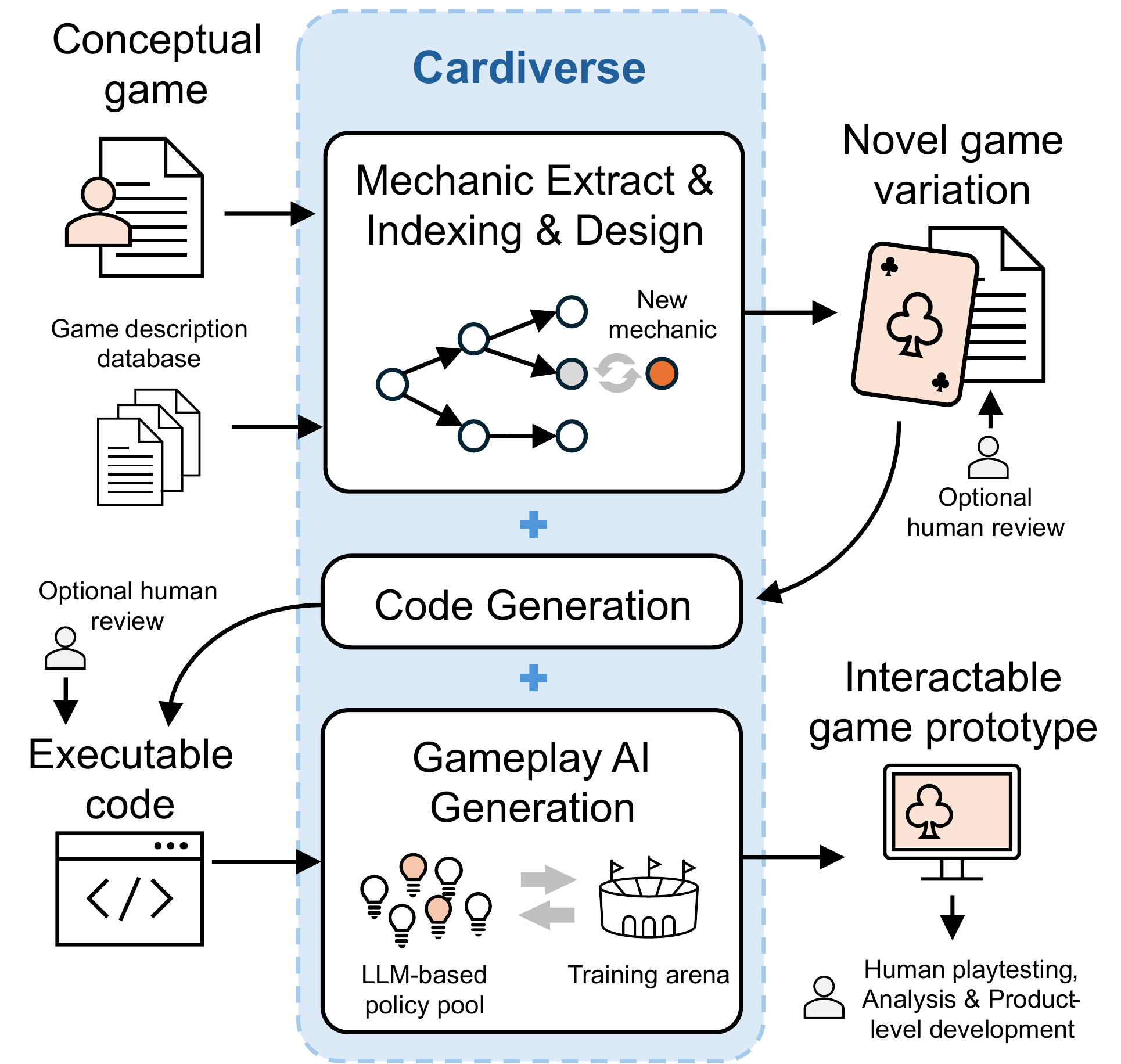}
  \caption{Our work integrates game mechanic design, code generation, and gameplay AI creation into an LLM-based framework. }\label{fig:concept}
\end{figure}

(1) \textbf{Novel variations in game mechanic}: 
Game mechanics, defined as the ``actions, behaviors and control mechanisms afforded to the player''~\cite{mda}, are iteratively varied during prototyping where novelty is encouraged.
Early studies created novel game variations by recompositing existing game mechanics that were extracted from game database~\cite{Guzdial_2022,pitako}. To further encourage novelty, recent work used LLMs to compose game mechanics from basic elements~\cite{charity2023conceptualgame}. However, it remains unclear how to encourage a more intentional novelty such that LLM outputs are neither too similar to existing games nor self-repeating.

(2) \textbf{Consistent game generation}: 
Translating game mechanics into consistent game code is critical for prototyping. To improve the consistency, prior work used LLM-generated I/O examples~\cite{liu2023is} to validate the generated code for short functions. This approach is not suitable for long programs with multiple rounds of interactions (such as card games), where the difficulty of generating correct I/O examples increases with the number of interactions.

(3) \textbf{Scalable evaluation with game AI}: 
Evaluating and refining game mechanic designs requires gameplay, which to be automated requires the support of AI agents with sufficient intelligence to explore game dynamics~\cite{Isaksen2015ExploringGS}. However, existing methods, ranging from reinforcement learning to LLM-based agents, come with significant drawbacks, including manual data formatting before training~\cite{zha_douzero_2021}, long training times~\cite{shinn_reflexion_2023} or costly inference~\cite{wang2023avalon}, which are impractical for rapid prototyping of diverse games.

To address these challenges, we propose \ourwork, an LLM-based card game prototyping pipeline that creates interactable card game variants given a base game description. It integrates game mechanic design, code generation, and gameplay AI creation into a cohesive framework, benefiting downstream playtestings and gameplay evaluations. The key contributions are:

\begin{enumerate}
    \item We propose an indexing method to represent games as game mechanic graphs. By clustering, summarizing, and designing new game mechanics as graph nodes, this approach informs LLM-based game mechanics design with a global understanding of existing databases. This enables the generation of novel game variations distinct from existing designs.
    \item We employ an LLM-based agent system that generates game code and iteratively reflects on its consistency by self-generated gameplay records. We transfer the reflection techniques on self-reported action trajectories, which are used in robotic task planning~\cite{innermonologue}, into game code generation. It improves the consistency between code and game mechanics, minimizing human labor in game prototyping.
    \item We introduce a scalable method for generating gameplay AI. Using a pool of action-value code functions produced by our framework, stepwise inclusion determines which functions to include based on win rates during self-play. Without calling LLMs in each game decision, it still achieves similar performance as LLM agents in prior work~\cite{yao_react_2023,shinn_reflexion_2023}. Therefore, it can conduct large-scale evaluations on new games generated by our pipeline, completing a cycle that enhances both creativity and efficiency in card game prototyping.
\end{enumerate}

\section{Related Work}

Large language models (LLMs) have demonstrated broad applicability across diverse design domains, with remarkable works ranging from video story ideation~\citep{xia-etal-2025-visiar,words2worldsLi2024} to architectural design~\citep{li2025archseek}. Within the realm of game design, prior research has explored their use in areas such as level creation~\citep{Todd_2023}, dialogue systems~\citep{akoury-etal-2023-framework}, and scene development~\citep{Hu_2024_Scenecraft,Kumaran_2023_scenecraft}. Focusing specifically on card games, this review narrows its scope to the non-visual components of gameplay, which constitute the core experiences of such games.

\subsection{Game Mechanics Design}

Assisting human in game mechanics design has been a long time topic, where the community made explorations on how to design mechanics that are both novel and feasible~\citep{cook2017angelina,automaticgame2008}. \emph{``P-creativity''}, which refers to the novel creation by reflecting on existing knowledge~\citep{BODEN1998creativity,guzdial2018creativeinventionbenchmark}, is one of the research focus. Early work on this topic produced new mechanics by searching a manually-defined parameter space \citep{automaticgame2008} or varying the combinations from a manually-defined set of basic types and rules~\citep{Chen_Guy_2020}. Later, the combinational variation method is enhanced by extracting a set of basic mechanics from video data or structured texts~\citep{sumner_mechanic_2024,Guzdial_2022,pitako}. Recently, as LLMs have shown its capabilities in creating content for interactive experience \citep{Todd_2023,words2worldsLi2024}, prior work further enhanced the pipeline by using LLMs to composite game mechanics from extracted basic elements~\citep{charity2023conceptualgame}.

Our work enhances the novel generations of LLM-based methods by retrieving game mechanic inspirations from a database. 
Resting on the cluster summarization techniques in \citet{edge2024localglobalgraphrag}, we group semantically similar mechanics across all games into clusters. Within each cluster, novel game mechanics are created from cluster summaries by LLMs. They are later sampled to enable a combinational variation on input games.

\subsection{Program synthesis in games}

Traditionally, many works focus on generating short code snippets in a specific domain such as matrix operations \cite{tfcoder} or list processing \cite{ellis_dreamcoder_2021}
. However, it is challenging to apply similar approach to game code generation, as it involves much larger search space, which does not suit iteration-based methods. 
Recently, large language models (LLM) have emerged as strong methods for program synthesis, where domain-specific fine-tuning~\cite{wu2024instructiondriven}, in-context learning 
\cite{Gao2023},
LLM agent systems
\cite{chatdev}
, and LLM pipelines \cite{agentless} are heavily studied. These methods have greatly improved code generation quality in many aspects, ranging from code snippets to real-world software engineering tasks~\cite{jimenez2024swebench}.

However, it remains unclear how to validate the consistency between user instructions and generated code for interactive code environments, since I/O examples~\cite{liu2023is} are hard to generate in these cases.

\subsection{Game Intelligence}
\label{sec:lr-gameai}

Creating AI systems capable of playing games intelligently has been a long-standing research focus. 
Early approaches relied on optimizing manually crafted game features~\cite{sturtevant_feature_2007} and then replaced by neural networks~\cite{schmid_student_2023,zha_douzero_2021,brown_superhuman_2019}. They show exceptional gameplay intelligence in some game domains.
But they require long training times, hyperparameter tweaking, and especially manual intervention to adapt game state representations to new games. Manual intervention hinders the scale with which we can evaluate the generated game variants, which makes them impractical for rapid prototyping.

Recent advancements in large language models (LLMs) offer a solution to this limitation as LLMs can adapt to diverse game inputs without case-specific customization. Prior work used LLMs as gameplay agents, wherein LLMs are invoked at every game turn \cite{yao_react_2023}, optionally supported by external storage for long-term memory or reflection~\cite{guo_suspicion-agent_2023,shinn_reflexion_2023,zhang2024agentpro}. 
Other studies combine LLMs with other neural networks by 
using LLMs to design reward functions \cite{ma_eureka_2023,baek2024chatpcg}, 
using neural networks to narrown down the search space for LLMs \cite{Yim2024guandan}, or training LLMs from scratch to predict game actions using real gameplay data \cite{schultz2024masteringboard}. 

To achieve scalable gameplay AI large-scale evaluations, we explore the following techniques that minimize the cost and latency in both construction and application stages: (1) we aims to get a gameplay policy in code \cite{liang_code_2023,light2024strategist} rather than leveraging LLMs in each game decision. (2) We do not optimize the policy by LLM-based reflections with gameplay data \cite{shinn_reflexion_2023}, as it can hardly be paralleled and introduce large noises. (3) We directly use win rate as optimization target rather than any fine-grained signals (such as reward by naive MCTS \citealt{light2024strategist}), as they may not be effective in certain games (e.g. deep game state trees).

\section{Method}

Our work composites an LLM-based card game prototyping pipeline by three components: In \ref{sec:game-logic} we leverage game mechanic graphs to enable novel game mechanic design, producing new game variations in text descriptions. In \ref{sec:game-code} we use the text descriptions from the previous step to generate code for the game. In \ref{sec:gameai}, gameplay AI is created using the generated code, thus enabling game evaluations for human designers.

\subsection{Game Mechanic Design}
\label{sec:game-logic}

The design problem is defined as below: given a card game database, where games are represented in text descriptions, we aim to create game variations that avoid high resemblance to the database. We tackle this by applying a variation instruction to LLMs, where we propose game mechanic replacements explicitly. Shown in Figure~\ref{fig:mechanics_pipeline}, our approach begins with an indexing process, where we extract mechanics as graphs from existing card game descriptions. The mechanics are subsequently clustered, and new mechanics are generated within each cluster using LLMs. During variation process, for a game to be variated, mechanics that frequently appear in the database are suggested to be replaced by newly-designed ones within the same cluster.

\subsubsection{Mechanics extraction}

\begin{figure*}[htbp]
  \centering
  \includegraphics[width=\textwidth]{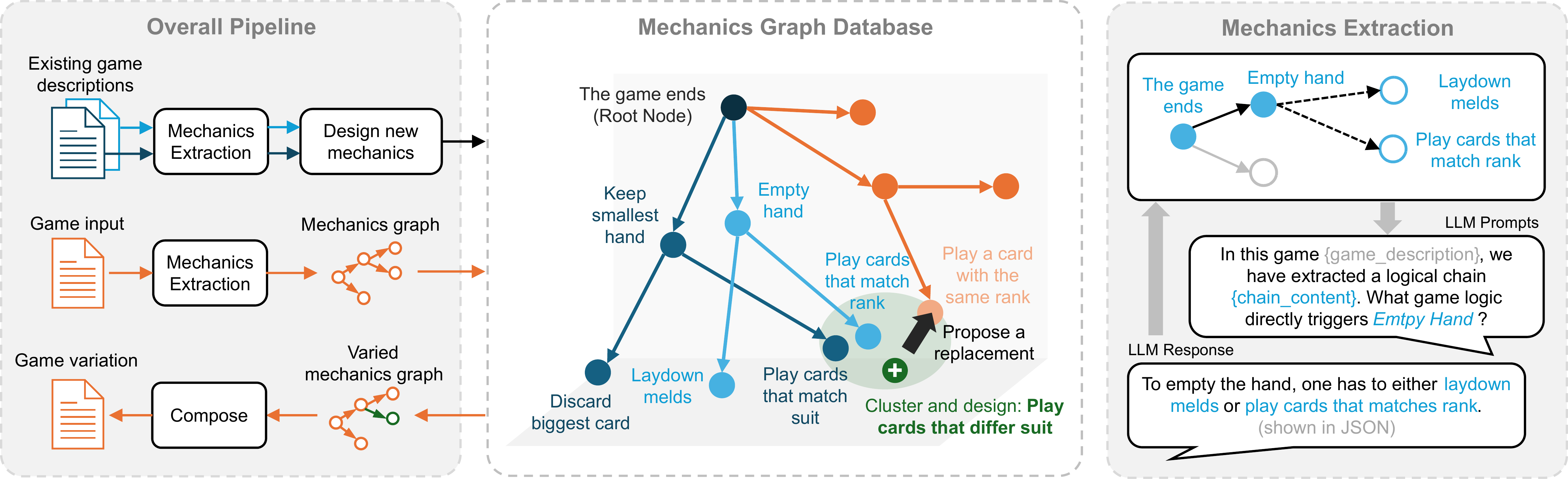}
  \caption{\textbf{Game mechanic design framework}. (Left) Overall pipeline consists of an indexing process (blue and black arrows) and a variation process (orange arrows). (Middle) Mechanic graphs shown in a text embedding space. The lower right corner showcase a new mechanics is created within a cluster (transparent green) and proposed as a replacement to a mechanic unit in the input game (light orange). (Right) Breath-first-search-like game mechanic extraction, where each expansion is done by LLM. }
  \label{fig:game_graph}
  \label{fig:mechanics_pipeline}
  \label{fig:mechanics_extract}
\end{figure*}

To encourage comprehensive and interconnected mechanics extraction from the text description, we propose a game mechanic graph representation. As illustrated in Figure \ref{fig:game_graph} (middle), game mechanics represented in short text phrases are stored as nodes, where all games share the root node ``the game ends''. The control flows between the nodes are represented as directed edges. The edge direction denotes the game effect contribution. For instance, in the game UNO, ``empty the hand'' is a downstream child of the root node because it directly leads to the game ending.

We adopt a backtracking approach to extract the graphs. Starting with the shared root node, we use LLMs to expand the search frontier like a breadth-first search. As illustrated in Figure~\ref{fig:mechanics_extract} (right), we input the game description and expansion history to the LLM (``the game ends'' $\rightarrow$ ``empty hand'') and instruct it to identify all mechanics that directly lead to the rear of the expansion history (``empty hand''). In the same way, we recursively expand all expanded nodes, where expansions are terminated by both the LLM's judgment and a manually set depth limit. It should be noted that while the mechanics are extracted in a tree search (thus the depths of nodes are recorded), we allow new nodes to be connected to any existing nodes so the extracted results may not be a tree.

\subsubsection{Mechanics Clustering and Designing}

We aggregate all mechanics from the game database and cluster them by semantic similarities and tree depth. Semantic similarity is measured using cosine similarity of text embeddings, with the \texttt{text-embedding-3-large} model from OpenAI serving as the embedding generator. A hierarchical clustering algorithm (UPGMA, see~\citealt{sokal1958statistical}) is employed, with a manually set similarity threshold of 0.4.

For clusters containing more than 3 mechanics, we input LLMs with all mechanic descriptions to summarize their shared themes and generate new mechanics adhering to these themes. 
To enhance diversity in the results, we vary the system prompts using techniques outlined in \citet{fernando_promptbreeder_2023}.
Smaller clusters are excluded from the summarization pipeline because they lack sufficient information for summarization tasks.

\subsubsection{Retrieval-augmented Game Variation}

The newly generated mechanics can be utilized to create game variations that differ significantly from existing database entries. Given a game description, we extract its game mechanic graph and map the mechanics to existing clusters. Mechanics from clusters with high occurrences in the database are identified as candidates for variation. Our framework then proposes replacing these frequently occurring mechanics with newly generated ones from the same cluster. An LLM applies this proposal to the game description, using a self-reflection process to resolve potential ambiguities or conflicts in the game mechanics. This approach ensures that the varied game remains coherent while introducing novel elements.

\subsection{Game Code Generation}
\label{sec:game-code}
\label{sec:codedraft}

Similar to \citet{agentless}, we adopt a workflow where LLM function calls are assembled into a predefined procedure. It receives a game mechanics description in natural language as input, and outputs the corresponding game code. 
The workflow first creates a game code without syntax /runtime error, and a structurized game description (detailed in \S~\ref{sup:init-code}). Then the validation process (Figure~\ref{fig:code-validate}) starts as below:

\begin{figure}[htbp]
  \centering
  \includegraphics[width=.48\textwidth]{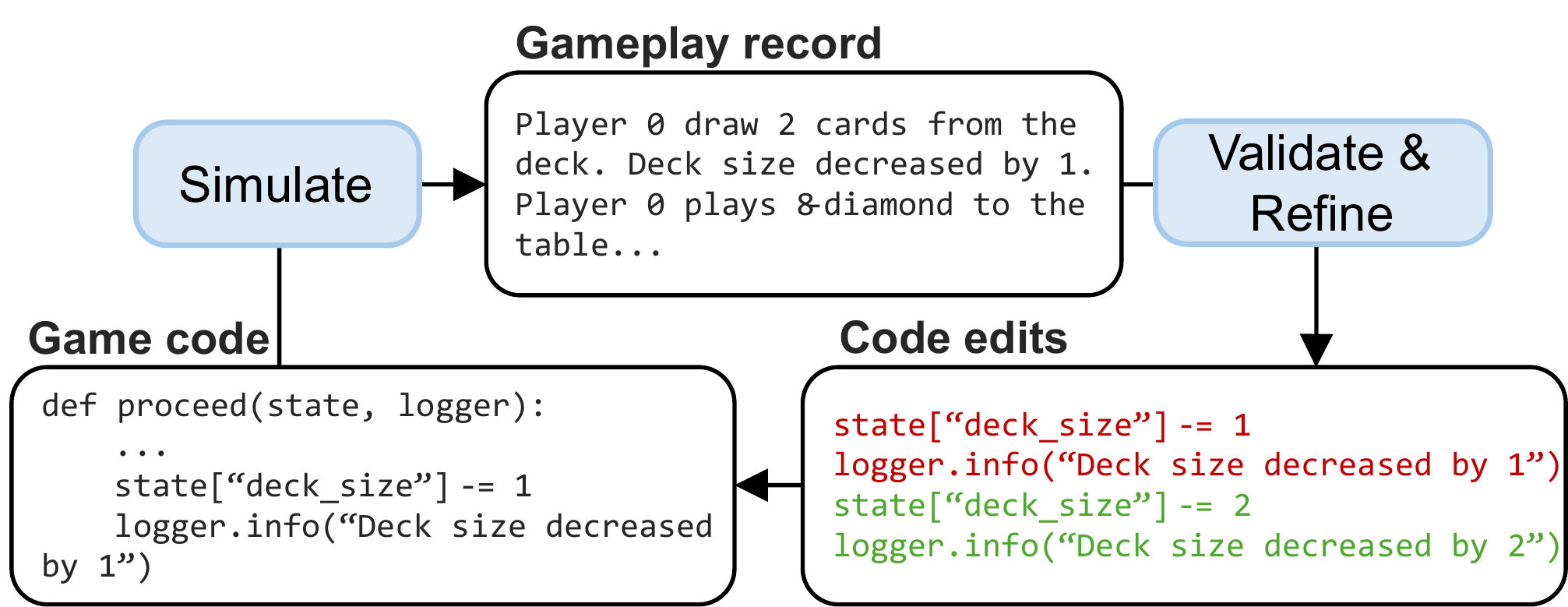}
  \caption{\textbf{Use gameplay records to validate code}. Gameplay records, generated by game code, are used to validate and refine the consistency of game code. }\label{fig:code-validate}
\end{figure}

\label{sec:validate}
We run the game code in a card game engine designed by us (detailed in \S~\ref{sup:game-engine}), sampling five gameplay records that are generated by the engine. The record includes the information from the last 6 game rounds and the game ending. In each round, the information includes the observation and the action of the current player, which are converted from the game state dictionary to natural language by a fixed parser. See \S~\ref{sup:code-validate} for gameplay record examples.

For each record, we send it to LLM with the structured game description, asking if the game play record violates any rule in the game description. If so, the violated rule, which is quoted by the LLM, is used as a query to a code snippet database.
This database retrieves related code snippets rather than entire code scripts. The retrieved code snippets act as the correct implementation of the rules that resemble the violated ones. It is fed to LLM with the previous violation analysis to refine the existing game code. After the refinement, the game code is sent back to the pipeline. This refinement loop continues until the maximum iteration is reached or the game code passes all validations from the records. See \S~\ref{sup:code-validate} for edit process examples.

\subsection{Gameplay AI Generation}
\label{sec:gameai}

\begin{figure*}[htbp]
  \centering
\includegraphics[width=\textwidth]{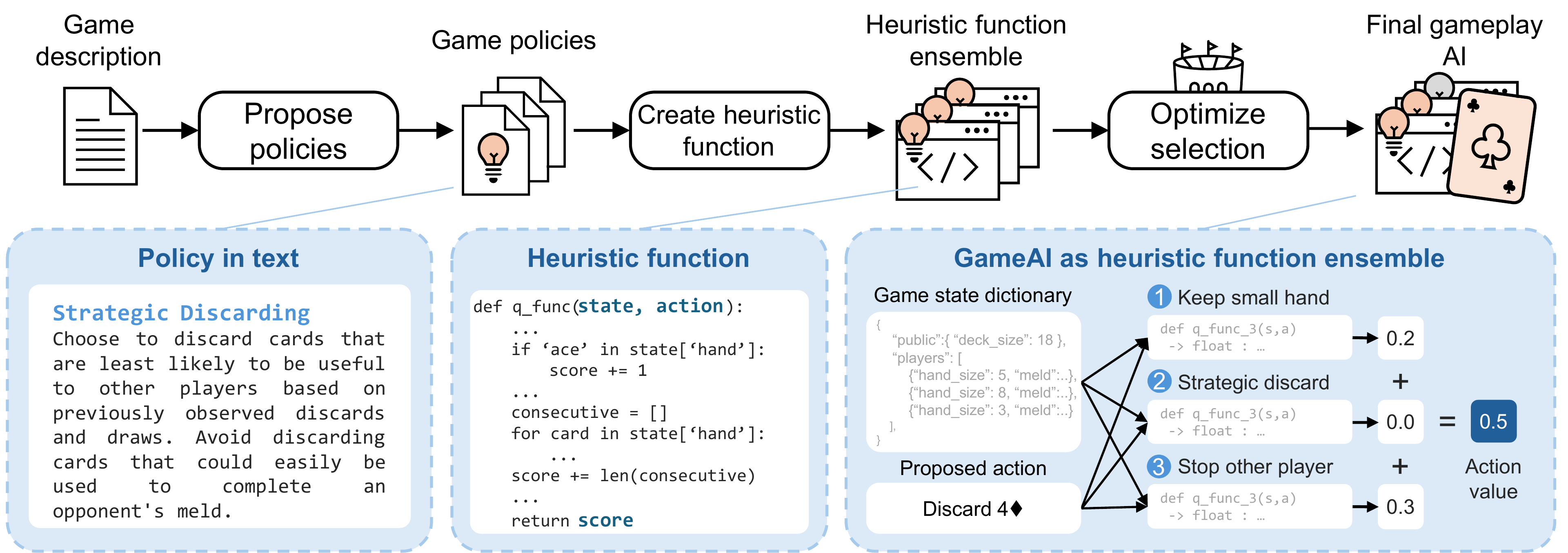}
  \caption{Gameplay AI generation pipeline.  }\label{fig:gameai_concept}
\end{figure*}

Our method constructs a gameplay agent that selects actions based on an ensemble of heuristic functions, each representing a distinct strategy derived from an LLM. These heuristics are implemented as Python functions that score candidate actions based on the current game state (similar to Q-functions in RL). The agent selects the action with the highest average score across all heuristics.

Formally, given the current game state $s_T$ and a discrete action space $A$, the gameplay agent selects the action $a_T$ that maximizes the average score assigned by a set of heuristic functions $\{Q^{\pi_1}, \dots, Q^{\pi_n}\}$, each derived from a distinct policy:
\begin{equation} 
    a_T = \text{argmax}_{a \in A} \frac{1}{n}\sum_{i=1}^n Q^{\pi_i}(s_T, a) 
    \label{eq: agent}
\end{equation}

Figure~\ref{fig:gameai_concept} illustrates the full pipeline. Starting from a natural language description of the game, we prompt an LLM to generate a diverse set of high-level policies in natural language. These are translated into Python-based heuristic functions that score actions. We then construct an ensemble from this set and iteratively optimize it through gameplay, using win rate evaluation to guide heuristic selection.

\subsubsection{Propose Policies}
\label{sec:propose-policy}

To bootstrap a diverse set of strategies, we prompt an LLM with a natural language description of the game and three distinct prompting methods. First, we instruct the LLM to generate natural language descriptions of possible ``strategies''. Second, we ask it to propose ``game state metrics that could be used for a reward function''. These two prompting methods are performed independently. Third, we prompt the LLM to revise each initially generated strategy by reflecting on its relationship to the previously proposed metrics. This reflection encourages policies that are grounded in in-game signals. See \S~\ref{sup:gameai-int-results} for examples.

In our experiments, we generate $n$ items for each prompting method, yielding a total of $3n$ distinct policies. 
We choose $n=4$ in our work (see \S~\ref{sup:sensitivity-policy-num} for further analysis).

\subsubsection{Construct Heuristic Function Ensemble}
\label{sec:ensemble}

Each policy above is translated into a Python function using the LLM. These heuristic functions accept the current game state and a candidate action as inputs (both represented as dictionaries) and return a scalar score representing the desirability of the action under the corresponding policy (see \S~\ref{sup:gameai-int-results} for examples). Code generation is automated, and each function is tested for syntax and runtime correctness. Functions that fail to execute without error are discarded.

To further enrich the ensemble, we generate a negative variant of each valid heuristic function by multiplying its output by -1. This simple transformation inverts the preference structure, producing $3n$ counter-policies that enhance the behavioral diversity of the ensemble. All $6n$ heuristic functions composite the full ensemble.

\subsubsection{Optimize Ensemble by Playing}

To select an effective subset from the full ensemble of heuristic functions, we use a step-wise inclusion strategy.
In the first iteration, each heuristic function $Q^{\pi_i}$ is evaluated individually by constructing a gameplay agent that uses only that function for action selection, as described in Equation~\ref{eq: agent}. The function with the highest win rate over 400 games against randomly behaving agents is selected and added to the current subset.

In each subsequent iteration, we evaluate every remaining candidate function (except the negative variations of the added heuristics) in combination with the current subset. A new agent is constructed for each candidate addition, and its win rate is again measured over 400 games against the same random players. The function that provides the greatest improvement in win rate is added to the subset. This process continues until no additional function improves performance.

The full optimization process consists of two phases. In the first phase, selection is performed against only random agents. In the second phase, optimization is repeated using a more competitive evaluation pool that includes both random agents and those selected in the first phase. The final subset from the second phase serves as the gameplay policy.

\section{Results}

We manually collected $106$ commonly seen card games from the web~\cite{bicyclecards2025howtoplay} as text descriptions, including 25 casino games (mostly poker), 27 trick-taking games, 11 rummy games, 10 solitaire-like games, and 33 others. These games are used to initalize the database for all three tasks (game design, code generation, gameplay AI).

Unless otherwise specified, we employed \texttt{gpt-4o-2024-08-06} as the LLM backbone and \texttt{text-embedding-3-large} from OpenAI as the text embedding model. All experiments were conducted using the default configuration of the OpenAI API. Specifically, we applied temperature sampling with a temperature of 1 and without nucleus (top-$p$) restrictions.

\subsection{Game Variations by Mechanics Design}

\begin{figure}[htbp]
  
    \begin{subfigure}[t]{.45\textwidth}
    \centering
        \includegraphics[width=.8\linewidth]{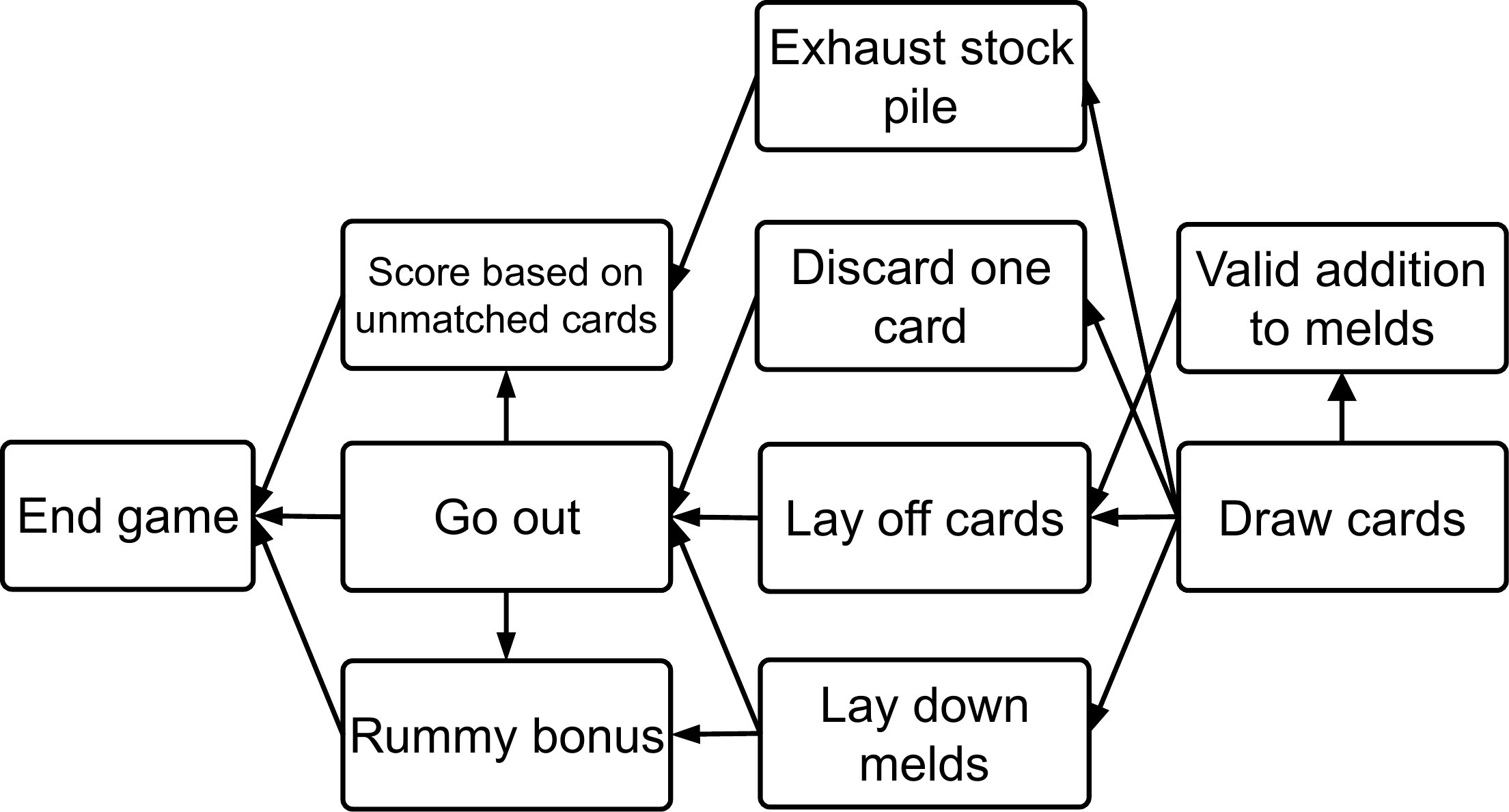}
        \caption{Boat House Rum}
        \label{fig:mechanic_1}
    \end{subfigure}
    \hfill
    \begin{subfigure}[t]{.45\textwidth}
    \centering
        \includegraphics[width=.8\linewidth]{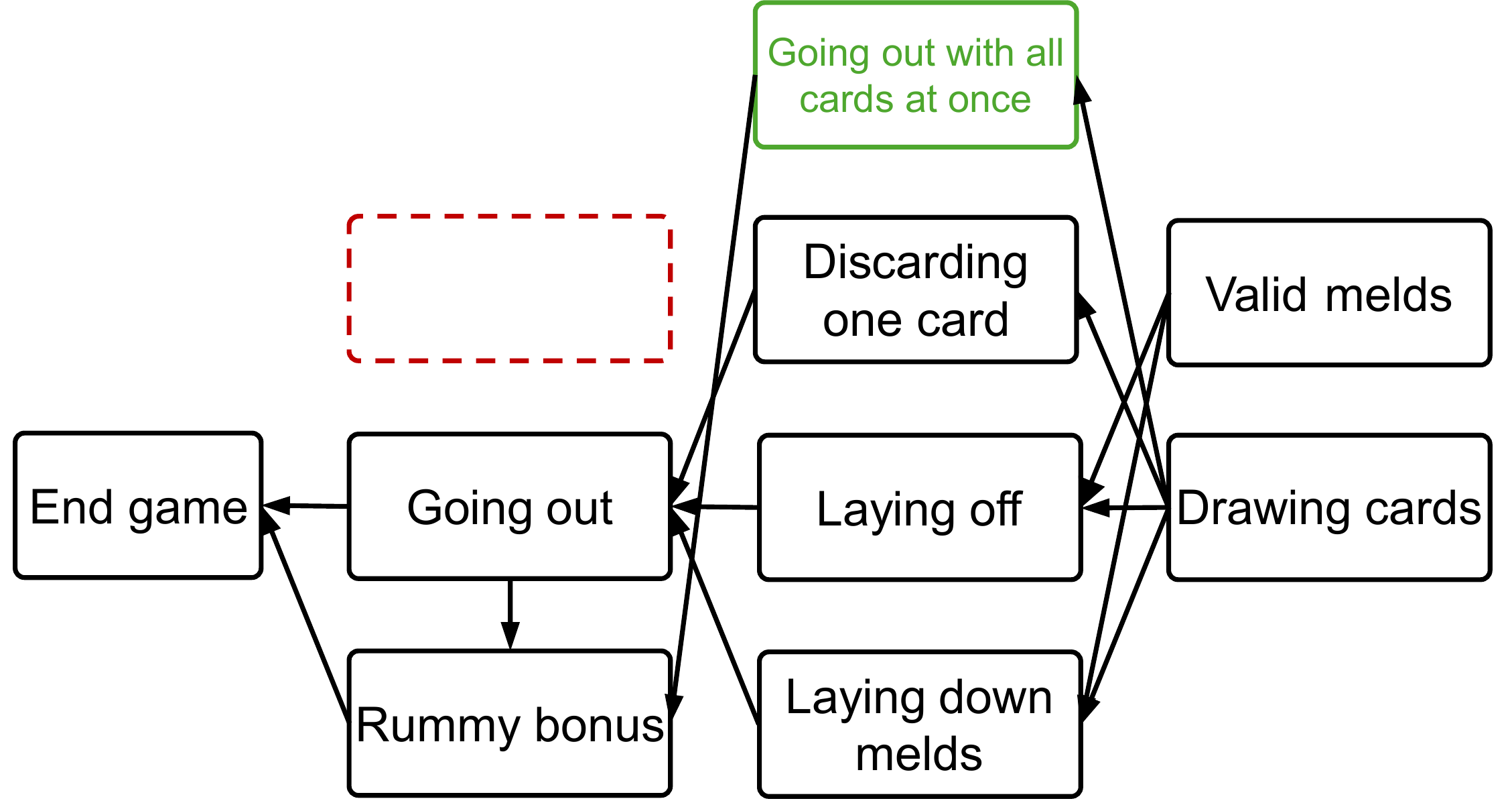}
        \caption{Rummy Rum}
        \label{fig:mechanic_2}
    \end{subfigure}
  \caption{Extracted game mechanics from two card games. The differences are colored in green (modification) and red (deletion).}
  \label{fig:extracted_graph}
\end{figure}

Our method starts with extracting game mechanics from all $106$ card games as graphs. Using our dataset, one game graph contains $13 \pm 5$ nodes and $20 \pm 10$ edges on average, depending on the complexity of game rules.
We show that robust and meaningful game mechanics are extracted, as the differences between extracted mechanic graphs aligns with human judgment. Figure~\ref{fig:extracted_graph} shows the graphs for two similar games (\emph{Boat House Rum} and \emph{Rummy Rum}), where we observe most of the nodes are consistent between two graphs except for the necessary differences. See \S~\ref{sec:ap-extracted} for more qualitative results.

Based on the mechanic extractions, our method can produce new mechanics for a cluster of similar mechanics. Here is an example where all mechanics that involve \emph{``win by emptying hands''} are clustered and new mechanics are designed as below. See \S~\ref{sec:ap-mechanic-clusters} for cluster examples and \S~\ref{sup:game-variants} for varied games as final products.

\begin{tcolorbox}[
breakable,
colframe=black, 
colback=white,
coltitle=black,
left=1mm, 
right=1mm, 
top=0mm, 
bottom=0mm, 
fonttitle=\small,
boxrule=0.2mm, 
]
\begin{lstlisting}[language=markdown]
[Get rid of all cards in hand]
A player wins the round by being the first to get rid of all their cards.

Newly-breeded instances:
- A player wins by strategically exchanging cards to end up with a singular, unique card that no other player holds.
- A player wins by collecting one of every type of card, achieving a complete set before others do.
- A player wins by maintaining an exact balance between cards acquired and cards played throughout the game.
- A player wins by successfully playing and then reacquiring their entire starting hand of cards, completing a full cycle.
- A player wins by discarding cards in a specific sequence to reveal a hidden pattern before anyone else does.
- A player wins by passing all their cards in an exact order to a partner, completing a successful relay of card exchange.
\end{lstlisting}
\end{tcolorbox}

We demonstrate that our method encourages novel game variations by measuring how its generated variations deviate from the original 106 games. We define Quantiled Max Similarity (QMS) as a metric. Given a base game $g$ and a variation method $M$, $QMS(M, g, p)$ represent the $p$th percentile of the closest cosine similarity scores between the variations of $g$ and every game in database $D$.
\begin{equation} 
    QMS(M, g, p) = \mathbf{Q}^p_{v \sim M(g)} \max_{d \in {D \setminus g}} \zeta(v) \cdot \zeta(d) 
    \label{eq: game_novel}
\end{equation}
where $\mathbf{Q}^p(\cdot)$ selects the quantile value at percentile $p$. $\zeta(\cdot)$ refers to the text embedding process. Due to limited computational resources, we choose $28$ games from $D$ as the evaluation set $G$. For each $g \in G$ we sample $33$ variations.
In this way, we can compare variation methods by examining their QMS distributions (through the interquartile range $p \in \{0.25,0.5,0.75\}$) across all games in $G$.

We compare our method against two alternatives: (1) Vanilla: the game variation is created by a fixed system prompt. Multiple variations are sampled by repeatedly querying the LLM. (2) PromptBreeder~\cite{fernando_promptbreeder_2023}: This work uses a predefined set of meta-prompts (e.g., ``Rewrite the following instruction to be more creative'') to automatically generate variations of a base system prompt. In our experiment, we applied this set of meta-prompts to our own system prompt to generate a diverse pool of variants, followed by manual filtering. Then we randomly pick the varied system prompts to generate game variations.

Table \ref{tab:game-sim} shows our method successfully penalizes game variations that are highly similar to database items. For variations that are in top 25\% or top 50\% on the similarity to database, our method outperforms PromptBreeder (abbreviated as \textit{``Prpt.Bdr.''}) by achieving a lower similarity score. PromptBreeder on the other hand, shows no significant difference to the Vanilla method.
For the rest of the variations (p=0.25), our method shows similar performance as PromptBreeder.

We also argue that the penalization does not make the variations to collapse to a limited number modes that leads to the loss of diversity. It is demonstrated by evaluating QMS within the generated variations, where a good method shall produces variations that differ from each other. To apply QMS within the variations, we replace the $D \setminus g$ in Equation \ref{eq: game_novel} with $M(g) \setminus v$.

As shown in the bottom section of Table \ref{tab:game-sim}, our method prevents high similarities than other methods (p=0.75). Our method offers fewer low-similarity variations (p=0.25) than PromptBreeder, since it offers more controllable process, where the variations are strictly directed by game mechanics replacements. 

\begin{table}[htbp]
    \centering
    \sisetup{separate-uncertainty= true, table-format=1.1(0), multi-part-units=single}
    \begin{tabular}{lS[table-column-width=1.4cm]S[table-column-width=1.4cm]S[table-column-width=1.4cm]}
        \hline
                            & \text{p=0.75} & \text{p=0.50}& \text{p=0.25} \\
        \hline
        \multicolumn{4}{l}{\textit{To database}}\\
        Vanilla        & 88.7\tiny{$\pm 5.8$}~$\dagger$ & 87.2 \tiny{$\pm 6.3$} & 83.8 \tiny{$\pm 8.2$} \\
        Prpt.Bdr.  & 88.3 \tiny{$\pm 4.9$}~$\dagger$& 85.8 \tiny{$\pm 4.5$} & 81.1 \tiny{$\pm 3.9$}~$\dagger$ \\
        Ours& \color{blue}87.4 \tiny{$\pm 4.5$}& \color{blue}84.8 \tiny{$\pm 4.3$} & \color{blue}80.6 \tiny{$\pm 3.7$}~$\dagger$\\
        \hline
        \multicolumn{4}{l}{\textit{Within}}\\
        Vanilla        & 95.4 \tiny{$\pm 0.8$} & 92.0 \tiny{$\pm 4.0$} & 86.7 \tiny{$\pm 5.9$}\\
        Prpt.Bdr.  & 91.0 \tiny{$\pm 2.1$} & 85.5 \tiny{$\pm 2.1$}~$\dagger$ & \color{blue}79.5 \tiny{$\pm 2.5$}\\
        Ours& \color{blue}89.6 \tiny{$\pm 2.4$} & \color{blue}85.4 \tiny{$\pm 2.8$}$~\dagger$ & 81.3 \tiny{$\pm 2.8$} \\
        \hline  
    \end{tabular}
    \caption{\textbf{Novelty performance between methods}. Means and standard deviations of Quantile Maximum Similarity ($\times 10^{-2}$) are reported in two scopes and three percentiles $p$. The first scope denotes the similarity between the variations and database. The second denotes the similarity within the variations. Pairs with $\dagger$ denote the data entries from the same column are NOT significantly different (p-value of paired t-test $\geq 0.05$)}
    \label{tab:game-sim}
\end{table}

Our generated game variants are also rated as subjectively more creative, with a user study indicating that participants found them strategically more engaging than those produced by prior methods (detailed in \S~\ref{sup:mechanics-user-study}).

\subsection{Consistent Code Generation}

We evaluate the pipeline on two sets of game descriptions: (1) 29 commonly-seen card games sampled from our collection of 106, and (2) 28 varied games generated using the method in \ref{sec:game-logic}, where each game in (1) is expanded into 33 variants and the variant with the lowest cosine similarity to the database is chosen. One game is not variated since its mechanics did not belong to a sufficiently large cluster for generating new mechanics.

The following metrics are used in our evaluation, where the second and third metric originate from \citet{chatdev}:

\paragraph{Generation Success (Succ)} This is a @pass3 metric using 10 tests per generated code. Given a game description, we generate code 3 times. Each script is executed 10 times with different random seeds for the card deck, using agents with random policies to maximize state coverage. If any script runs error-free in all 10 trials, the task is marked as successful. Table \ref{tab:codegen} shows results as “successful tasks/total tasks.”

\paragraph{Executability (Exec)} Similar to \emph{Succ}, we run each game code 100 times with random seeds. A run is successful if it completes without syntax or runtime errors. We report the average success rate across all codes that passed \emph{Succ}.

\paragraph{Embedding Consistency (ECon)} Semantic embedding similarity between the code and the description of the game. Here we used structurized description detailed in Section~\ref{sec:codedraft}. While this metric may lack task-specific precision, we include it for consistency with prior work (\emph{e.g.}, \citealt{chatdev}) to facilitate cross-study comparison.

\paragraph{Play Record Consistency (PCon)} We sample 10 game play records using randomized decks and policies to maximize state diversity. Each record is evaluated by an LLM to score the consistency (scaled 1 to 10) between actual game play record and game description, which applies a similar implementation as \ref{sec:validate}. We use \texttt{claude-3-5-sonnet-20241022} here to avoid the blind spots of one single LLM source. See Appendix~\ref{sup:pcon-eval} for the details of game play record. We also test the effectiveness of PCon in Appendix~\ref{sup:pcon-eval}.

Our method is compared against its ablated version, where LLM-based validation is skipped after the repeated debugging (abbreviated as ``-val''). We add the second ablated version, where validation is skipped and the LLM backbone is replaced by OpenAI \texttt{o1-preview} in code initialization (``-val$\oplus$'').

Table \ref{tab:codegen} demonstrates a robust code generation using our method, as the mean executability is over $90\%$. 
Also, our method improves the code alignment to the game description, as play record consistency metrics in both game groups (common and varied) are higher than its ablated version. While a stronger model results in higher generation success rate, for successfully-created games, our method lifts the consistency of a weaker model to a strong one, indicating its potential to substitute stronger backbone models when they are not available.

\begin{table}[htbp]
    
    \sisetup{separate-uncertainty= true, table-format=1.2(2), multi-part-units=single}

    \begin{subtable}[htbp]{.48\textwidth}
    \centering
    \begin{tabular}{lcS[table-format=2.1(0)]S[table-format=1.2(0)]S[table-format=0.2(0),table-column-width=1.6cm]}
        \hline
                  & \text{Succ$\uparrow$} & \text{Exec$\uparrow$} & \text{ECon$\uparrow$}& \text{PCon$\uparrow$}\\
        \hline
        Ours    & \text{27/29} & \color{blue}100.0\tiny{$\pm 0$} & 0.55 \tiny{$\pm 0.04$} & \color{blue}8.99\tiny{$\pm 1.56$} \\
        \text{-val} & \text{27/29} & 99.9\tiny{$\pm 3.9$} & \color{blue}0.56 \tiny{$\pm 0.03$}~$\dagger$ & 7.69 \tiny{$\pm 1.96$}~$*$ \\
        -val$\oplus$& \color{blue}29\text{/29} & 96.3\tiny{$\pm 9.4$} & 0.56 \tiny{$\pm 0.04$} & 8.90 \tiny{$\pm 1.74$}~$\dagger$ \\
        \hline
    \end{tabular}
    \caption{Common Games}
    \end{subtable}
    \hfill
    \begin{subtable}[htbp]{.48\textwidth}
    \centering
    \begin{tabular}{lcS[table-format=2.1(0)]S[table-format=1.2(0)]S[table-format=0.2(0),table-column-width=1.6cm]}
        \hline
                  & \text{Succ$\uparrow$} & \text{Exec$\uparrow$} & \text{ECon$\uparrow$}& \text{PCon$\uparrow$} \\
        \hline
        Ours    & 25/28  & \color{blue}100.0\tiny{$\pm0$} & 0.52 \tiny{$\pm 0.05$}  & \color{blue}9.20 \tiny{$\pm 1.08$}\\
        \text{-val} &  25/28  & 98.3\tiny{$\pm4.6$} & 0.52 \tiny{$\pm 0.04$} & 8.66 \tiny{$\pm 1.60$}\\
        -val$\oplus$&  \color{blue}28/28  & 99.3\tiny{$\pm2.0$}   & \color{blue}0.54 \tiny{$\pm 0.05$}& 9.08 \tiny{$\pm 1.29$}~$\dagger$\\
        \hline
    \end{tabular}
    \caption{Varied Games}
    \end{subtable}
    
    \caption{\textbf{Code generation performance between methods}. Mean values of the metrics across all game instances are reported. The standard deviation is shown after $\pm$. $*$ represents the metric are significantly different (p-value $\leq 0.05$ in paired t-test) from our method, while $\dagger$ denotes the metric are unlikely to be different (p-value $\geq 0.7$) from our method.}
    \label{tab:codegen}
\end{table}

\subsection{Gameplay AI Generation}

After a manual verification on all generated games (both common and varied) in the previous task, some games are excluded from the test set for gameplay AI, as they are either purely luck-oriented games with no strategies involved, or they are not completely consistent with the corresponding game description. Ultimately, we selected $13$ common games and $6$ varied games (listed in \S~\ref{sup:gameai}) for gameplay AI evaluation.

We benchmarked our method against Chain-of-Thought \cite{CoT}, ReAct \cite{yao_react_2023}, Reflexion \cite{shinn_reflexion_2023}, and Agent-Pro noted as BeliefAgent \cite{zhang2024agentpro} (detailed in \S~\ref{sup:game-ai-eval}). Non-LLM methods are not included since they are not suitable for the game prototyping context (see \S~\ref{sec:intro} and \S~\ref{sec:lr-gameai}).

The win rate performance of a method is assessed by its advantage over competing policies. Specifically, we measure how effectively policy \( p_2 \) outperforms policy \( p_1 \) under identical contexts. In games with \( n \) players, the first \( n-1 \) players act as defense, and the last player as the attacker. The win rate of an attacker using \( p_1 \) against defenders using \( p_2 \) is denoted as \( \omega(p_2, p_1) \). The advantage of \( p_1 \) over \( p_2 \) is defined as:
\begin{equation} 
    A(p_1, p_2) = \omega(p_2, p_1) - \omega(p_2, p_2)
\end{equation}

Table \ref{tab:gameai-adv} shows that our agent achieves the highest win rate advantage ($p_2$ = random agent), with significant differences over all the opponent agents in common games. For varied games, our agent not only achieves the highest average win rate but also is the only agent that can beat the random agents. We also show its performance against human-designed heuristics in \S~\ref{sup:gameai-heuristics} and ablation experiments in \S~\ref{sup:gameai-ablation}.

\begin{table}[ht]
    \centering
    \sisetup{separate-uncertainty= true, table-format=2.1(1), multi-part-units=single}
    \begin{tabular}{@{}lccc@{}}
        \hline
               & \text{Common} & \text{Varied} & \text{All} \\
        \hline
        CoT    & 1.4 \tiny{$\pm 15.8$} & -4.8 \tiny{$\pm 13.4$}  & -0.5 \tiny{$\pm 15.0$} \\
        ReAct  & 2.4 \tiny{$\pm 16.4$} & -4.5 \tiny{$\pm 9.9$} & 0.2 \tiny{$\pm 14.7$} \\
        Reflexion & 1.4 \tiny{$\pm 18.4$} & -3.2 \tiny{$\pm 11.7 $} & 0.0 \tiny{$\pm 16.4$} \\
        Belief & 4.2 \tiny{$\pm 18.2$} & -1.8 \tiny{$\pm 18.0$} & 2.3 \tiny{$\pm 17.9$} \\
        Ours &  \color{blue} 14.6 \tiny{$\pm 19.3$}~$*$ & \color{blue}  19.7 \tiny{$\pm 30.5 $} &\color{blue} 16.3 \tiny{$\pm 22.6$}~$*$\\
        \hline
    \end{tabular}
    \caption{\textbf{Game AI win rate advantage between methods}. Means and standard deviations (shown after $\pm$) of win rate advantages ($\times 10^{2}$) over random agents across all game instances. $*$ represents our metrics are significantly different from those of prior work (p-value $\leq 0.05$ in paired t-test).}
    \label{tab:gameai-adv}
\end{table}

\begin{figure}[ht]
    
  \centering
  \includegraphics[width=.45\textwidth]{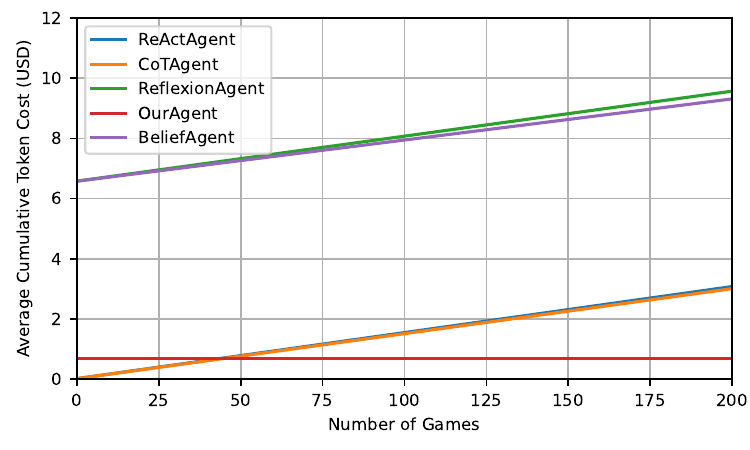}
  \caption{\textbf{Gameplay AI Token Cost}. Total LLM cost as a function of number of game runs,  averaged across 19 games. The cost after 50th run is estimated by linear extrapolations.
  }
  \label{fig:token_cost}

\end{figure}

In the cost analysis, we used OpenAI's API token pricing as a reference. We take the average per-game token cost across all 19 games as the metric here. As our agent only requires one-time usage of tokens during the generation phase, our total cost becomes lower than prior work in just 50 game runs (see Figure~\ref{fig:token_cost}), which suits better for large-scale evaluations in game prototyping. Refelxion and Belief agent have a significantly higher cost during training phase.

\section{Conclusion}
This work introduces an LLM-based pipeline for card game prototyping. It integrates three key components: (1) a graph-based representation of game mechanics that facilitates the extraction, clustering, and generation of novel game mechanics; (2) a game code generation system that aligns code with text description through iterative gameplay validation; and (3) a scalable gameplay AI that generates ensembles of action-value functions optimized through self-play.
It provides designers with interactable game variants at minimum implementation cost, enabling faster card game prototyping cycles for human designers. 
Future work may consider versatile game variation methodologies, broader game genres, or more powerful code generation and game AI.

\section*{Use of Scientific Artifacts}

Our use of card game data from~\citet{bicyclecards2025howtoplay} falls within the scope of fair use as defined under 17 U.S.C. \S 107.
\section*{Limitations}

We outline the limitations and potential future work in the following parts:

\paragraph{Game variations}
Our work proposes a method to create new game mechanics from existing game descriptions, and further showcases how the new mechanics help creating game variations by replacing one mechanic in the game input. 
Our decision to focus on relatively simple variation methods was deliberate. We argue that a foundational challenge, one that has been underexplored in prior work, is how to systematically measure and account for novelty in game variations. We believe that this issue must be addressed before more complex structural changes can be meaningfully pursued.

To this end, we adopt a strategy of making small, controlled modifications to existing games. This allows us to rigorously investigate how leveraging a structured representation (in our case, a graph database) can support the generation of novel yet coherent variations.

However, new mechanics can also be integrated in other combinational methods (shown in \citealt{Guzdial_2022}), which can be solved in future work.

Also, while our framework supports creating variations from human input, our current implementation expects human inputs to match the detail level of our dataset examples. Handling more colloquial or incomplete human inputs remains a promising direction for future research.

Furthermore, our work relies on LLM reflections to enhance the coherence between newly-added mechanics and existing ones. This approach reduces the human labor in solving the potential conflicts, as prior work requires human-computer cooperation~\cite{pitako}. However, future work could explore whether the extracted mechanics graph can help the advanced reasoning on potential game mechanics conflicts.

\paragraph{Scale mechanics design to other game genres}
Our work can be applied to the logical dependencies in other game genres. For example, by changing the node content from game mechanics to atomic story beats, we could apply it to the nonlinear/branching narratives in story-rich games. Similarly, we can also adapt to the dependencies of puzzles/tasks in adventure puzzle games.
But in broader scenarios, we expect a larger number of game mechanics for some cases. While our current clustering method can be controlled by semantic similarity, its trivial implementation requires $\mathcal{O}(n^3)$ time complexity. Switching to Gaussian Mixture Models ($\mathcal{O}(n)$ time) may be a better scalable solution, though it lacks intuitive clustering control.

\paragraph{Game code generation}
In game code generation, the current validation process is not efficient enough, making its token cost close to that of adavanced model (such as o1). Future work may aim to lower the token use by other agent pipeline designs.

Also, code generation is sensitive to the instruction-following ability of the LLM. We found that weaker models struggle to produce code aligned with game descriptions. In contrast, other tasks in our framework (\emph{e.g.} graph extraction and mechanic summarization) are less affected by the LLM choice.

\paragraph{Gameplay AI}
Currently, our game AI does not explicitly consider other players' intention, which is a commonly-used component in prior work \cite{zhang2024agentpro,guo_suspicion-agent_2023}. Besides, it could also be promising to explore whether the reasoning results from LLM agents (such as Reflexion \citealt{shinn_reflexion_2023}) can be distilled to our policy-code-based results.
\bibliography{main.bib}
\newpage
\appendix

\section*{Appendix for ``Cardiverse: Harnessing LLMs for Novel Card Game Prototyping''}


\startcontents[mytoc] 

\printcontents[mytoc] 
{l}{1} 
{\setcounter{tocdepth}{2}} 

\vspace{10mm}

\section{Game Mechanics}
\subsection{Extracted Game Mechanics}
\label{sec:ap-extracted}
\begin{figure}[htbp]
  \centering
    \includegraphics[width=.45\textwidth]{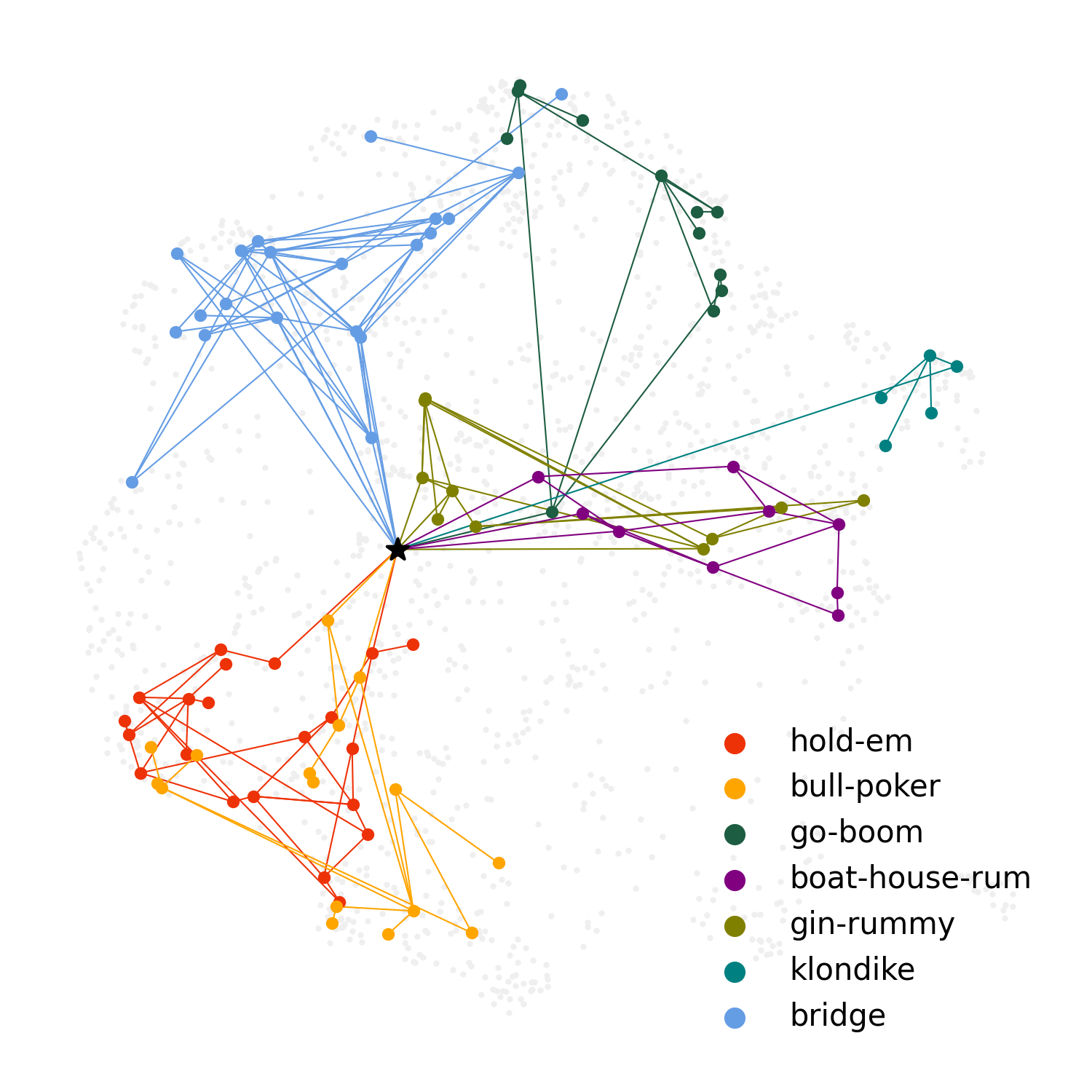}
  \caption{\textbf{Extracted mechanics in an embedding space}. Mechanics are distributed in a 2D UMAP projection of their text embedding space, colored by its game ID. Selected card games shown as mechanic graphs, where the shared root node is displayed in black star mark. }
  \label{fig:mechanic_graphs}
\end{figure}

Figure \ref{fig:mechanic_graphs} shows a more macroscopic comparison among multiple games, where the mechanic graphs of two poker games (\textit{Bull Poker} and \textit{Holdem}) are closer to each other than a card game from a different genre (\textit{Go Boom}).  

\begin{figure}[htbp]
  \centering
    \includegraphics[width=.45\textwidth]{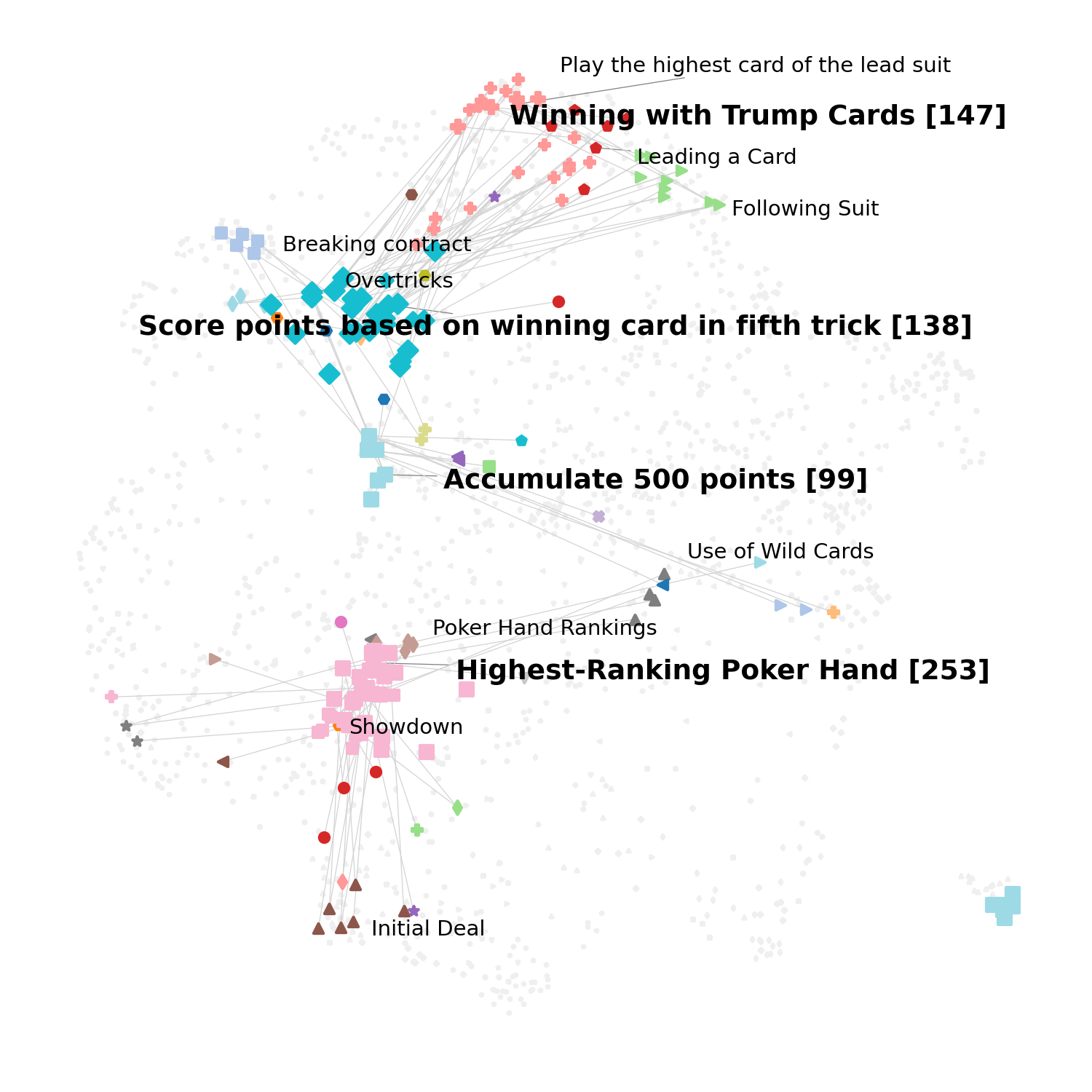}
  \caption{\textbf{Extracted mechanic dependencies}. Mechanics are distributed in a 2D UMAP projection of a text embedding space, colored by its cluster ID. Starting from selected mechanic clusters (in bold text), their dependencies on supporting mechanics (in normal text) are drawn in lines. }
  \label{fig:mechanic_dependencies}
\end{figure}

Also, our method reveals the dependency between game mechanics. In Figure \ref{fig:mechanic_dependencies}, the mechanics ``Highest-ranking poker hand", which frequently appears in poker games, is associated with relevant mechanics such as ``showdown" and ``poker hand rankings". 

\begin{figure}[htbp]
  \centering
    \begin{subfigure}[htbp]{.45\textwidth}
        \centering
        \includegraphics[width=\linewidth]{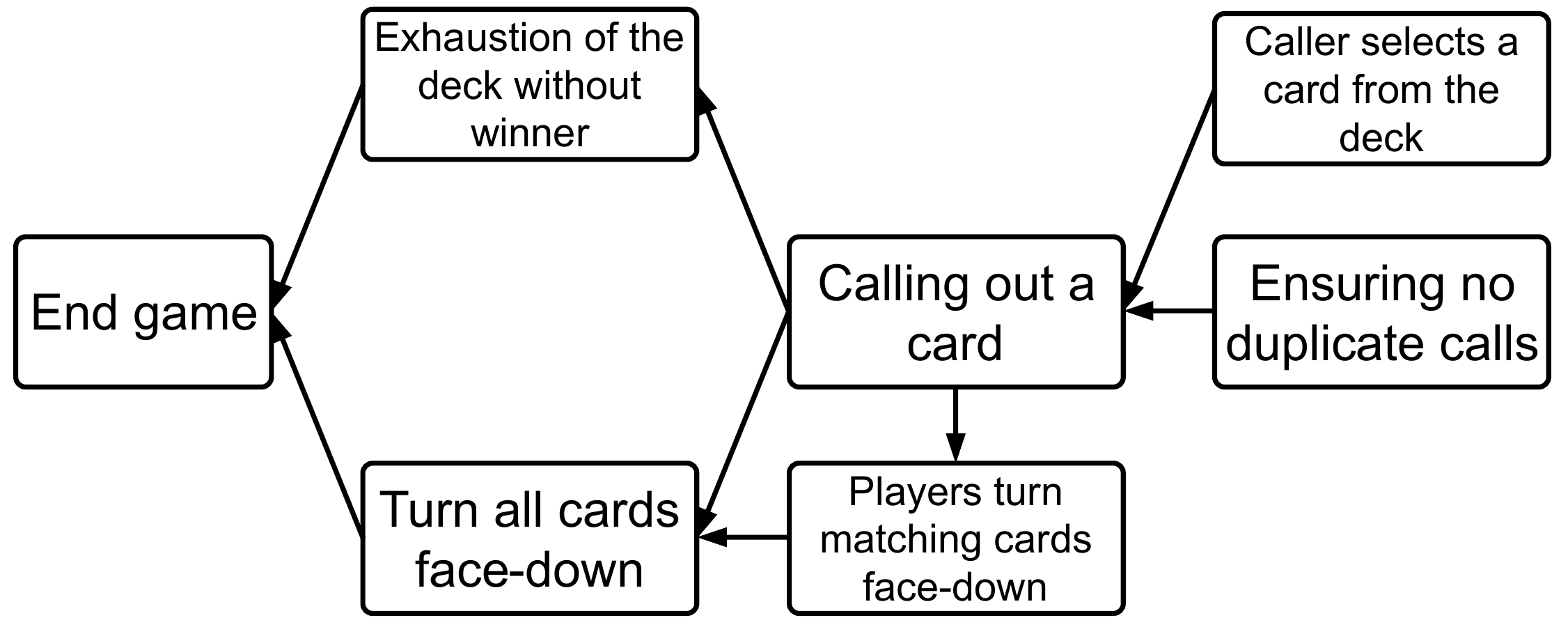}
        \caption{Bingo}
        \label{fig:more_mechanic_1}
    \end{subfigure}
    \hfill
    \begin{subfigure}[htbp]{.45\textwidth}
        \centering
        \includegraphics[width=\linewidth]{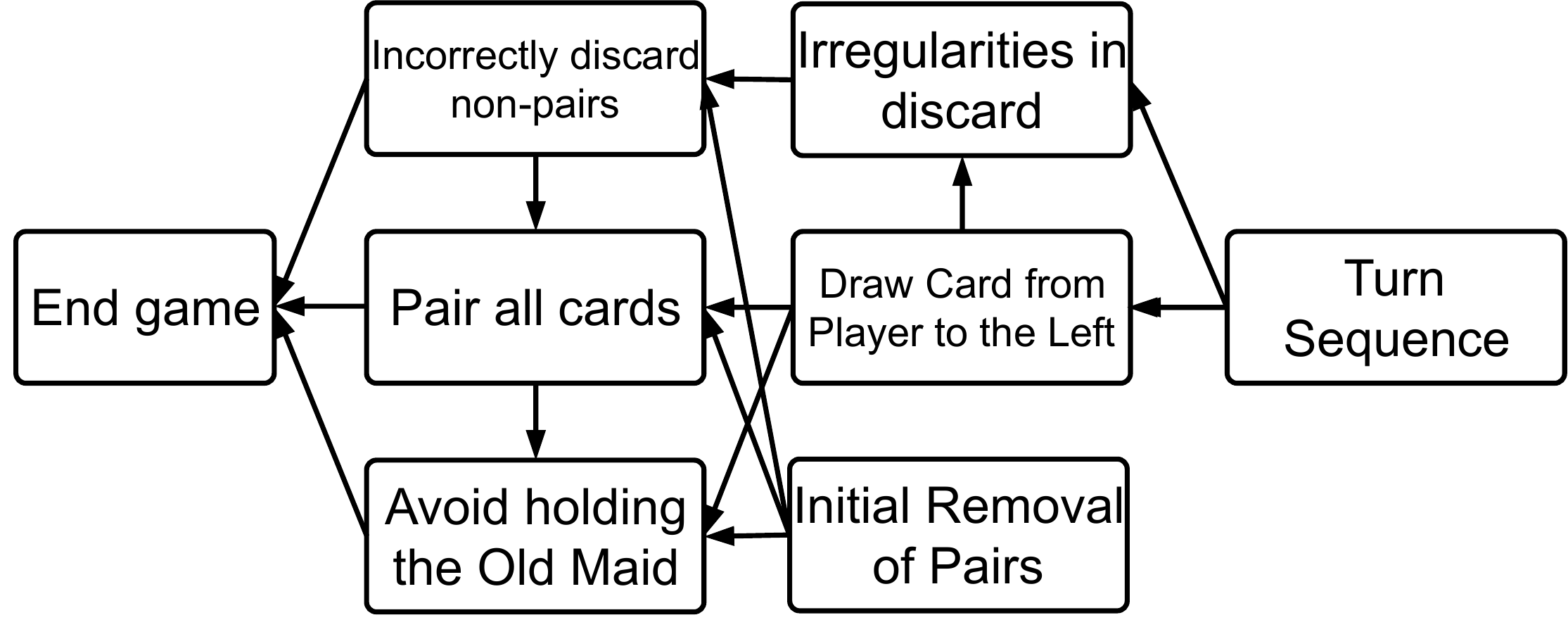}
        \caption{Old Maid}
        \label{fig:more_mechanic_2}
    \end{subfigure}
    \hfill
    \begin{subfigure}[htbp]{.45\textwidth}
    \centering
        \includegraphics[width=\linewidth]{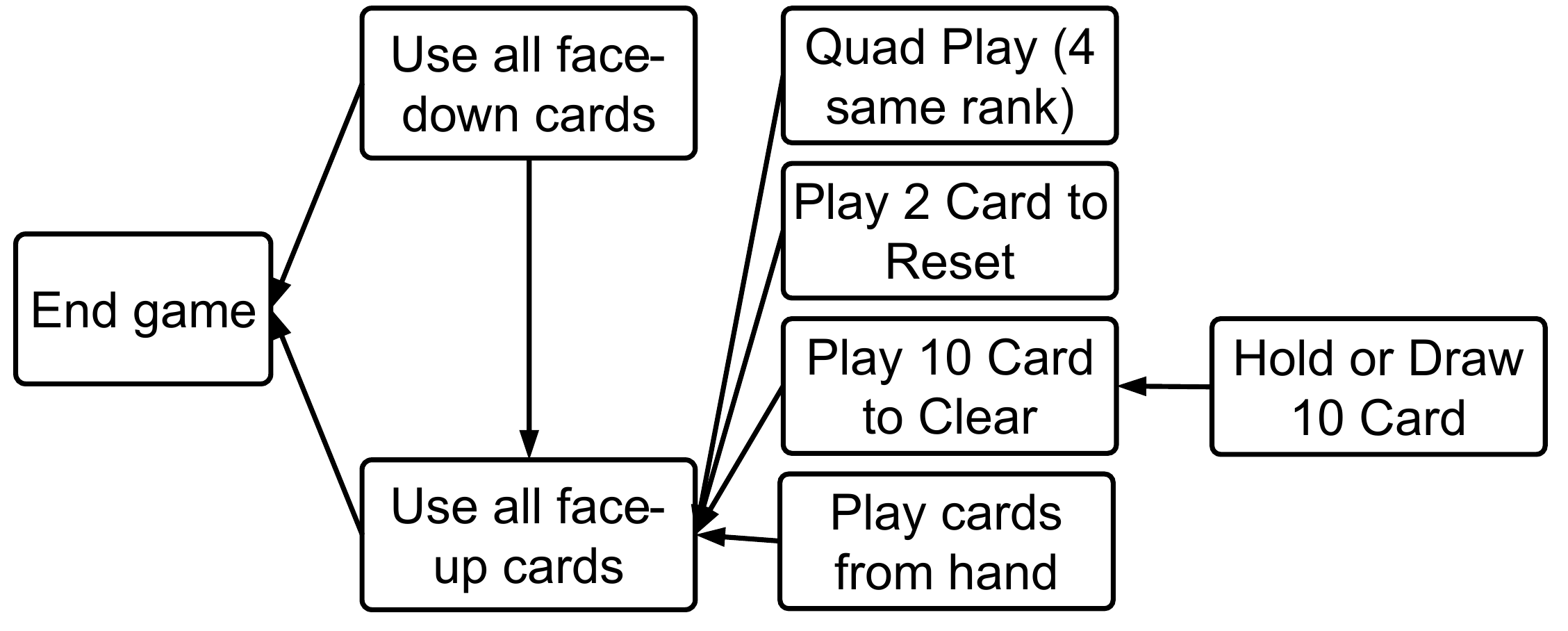}
        \caption{Palace}
        \label{fig:more_mechanic_3}
    \end{subfigure}
  \caption{More extracted game mechanics from game descriptions.}
  \label{fig:more_extracted_graphs}
\end{figure}

See Figure~\ref{fig:more_extracted_graphs} for more qualitative results of extracted game mechanics.

\subsection{Mechanics Cluster}
\label{sec:ap-mechanic-clusters}

We present two examples of mechanic clusters. The first is drawn from clusters whose mechanics lie one level away from the root node in the game mechanic graph, representing more fundamental mechanics. The second is drawn from clusters two levels away from the root, corresponding to more atomic mechanics. Each example is described with the following components:
\begin{itemize}
\item \textbf{Abstraction}: The overarching theme that unifies all mechanic instances within the cluster, as summarized by the LLM agent in our pipeline.
\item \textbf{Variation}: The design principles for generating new mechanics within the cluster, also proposed by our LLM agent.
\item \textbf{Instances}: The complete set of mechanic instances, drawn either from an existing game database or generated by our method (with newly designed instances marked by ``New'' in curly brackets).
\end{itemize}

\begin{tcolorbox}[
breakable,
title=Mechanic Cluster Example 1,  
colframe=darkgray, colback=white,coltitle=white,
left=1mm,right=1mm,top=0mm, bottom=0mm, fonttitle=\small, boxrule=0.2mm, 
]
\begin{lstlisting}[language=markdown]
Concept 100: Scoring Mechanics Game

**Abstraction**
The game's objective revolves around achieving or maintaining specific point thresholds or targets to determine a winner.

**Variation**
The instances vary in the target winning conditions and mechanics, such as reaching a specific positive point threshold (e.g., 10, 30, 51, 100, 121) to win versus strategies focused on having the lowest score to win at the game's end when another threshold (e.g., 100 points) is hit. Some instances award bonus points for certain achievements, while others penalize players with high scores or require low penalty scores to win.

**Instances**
### Instance 0: Receive Bonus Points for Winning
An additional 10 points are awarded to the player who has the highest cumulative score at the end of the game.

### Instance 1: Reach a score of 30 or more points
A team wins the game by reaching a score of 30 or more points.

### Instance 2: Score lowest at the end of the game
The game ends when a player reaches 100 cumulative points. The player with the lowest score is declared the overall winner.

### Instance 3: Accumulate 100 points
The first player to reach 100 points across multiple hands wins the game.

### Instance 4: Lowest score upon end of game
When the game ends by a player reaching 100 points, the player with the lowest score wins the game.

### Instance 5: Score 10 or more points
The game ends when a player scores 10 or more points.

...(remaining instances are not shown here)...


### Balanced Trade-Off Scoring (New)
Players navigate the game by accumulating points, with the goal of ending exactly between 50 and 70 points. Crossing this range results in starting the next round with a penalty of 10 points. The first to finish a round within this range for three consecutive rounds declared the winner.

### Pivot Point Strategy (New)
Players must reach precisely 50 points to trigger a special round where they can choose to either lock their score or risk doubling it with a final challenge. The first player to successfully double their score beyond 100 points without exceeding 120 points wins the game.

### Score Flip Challenge (New)
Players aim to score exactly 60 points. Upon reaching this exact score, their points flip, causing the player to adopt a defensive strategy to end the game. The player wins by maintaining the least fluctuation in score after the flip until the game ends.

### Controlled Climb (New)
Players race to reach exactly 75 points. Upon reaching 75 points, they must complete one additional round trying not to exceed their score from any previous rounds. The winner is the player who reaches 75 points first and maintains the lowest maximum score in the additional round.

### Dynamic Range Mastery (New)
Players aim to end with scores between evolving target ranges that shift every round (e.g., 20-40 in Round 1, 50-70 in Round 2, etc.). The player who consistently lands within each target range without going over wins the game after five rounds.

### Threshold Tempo (New)
Players strategize to reach multiple designated scoring targets (e.g., 30, 60, 90) within a set number of rounds. Achieving a target grants a strategic advantage, whereas missing it induces a point freeze. The winner is the first player to hit all targets exactly and complete the final round with no point freeze active.
\end{lstlisting}
\end{tcolorbox}

\begin{tcolorbox}[
breakable,
title=Mechanic Cluster Example 2,  
colframe=darkgray, colback=white,coltitle=white,
left=1mm,right=1mm,top=0mm, bottom=0mm, fonttitle=\small, boxrule=0.2mm, 
]
\begin{lstlisting}[language=markdown]
Concept 145: Follow Suit Card Game

**Abstraction**
The core concept revolves around players trying to follow the suit of the lead card if possible. If they cannot follow the suit, they are often required to draw from the stock or play any card depending on additional rules.

**Variation**
While the central theme requires players to follow the lead suit, variations include drawing from the stock when unable to follow, having options to play any card initially or using a trump card, and various specific rule-based outcomes like collecting tricks or adding cards to hand if unable to follow the suit.

**Instances**
### Instance 0: Follow Suit
Players must play a card of the same suit as the lead card if possible; they may play a trump card or any card if unable to follow suit.

### Instance 1: Have a Card of the Led Suit
Players must have a card of the same suit as the lead card to follow the suit.

### Instance 2: Following the suit led
Players must follow the suit of the initially led card if they have any in hand.

### Instance 3: Having a legal card to play
A player can play a card if they have one that matches the suit or rank of the lead card.

### Instance 4: Unable to play leads to drawing from stock
If a player does not have a card matching the suit or rank, they must draw from the stock pile.

### Instance 5: No card in hand matches the lead suit or rank
A player is unable to play a card from their hand if none of their cards match the suit or rank of the lead card.


...(remaining instances are not shown here)...


### Cascade Suit Challenge (New)
Players must follow the suit of the lead card or play a card of the same rank. If neither is possible, they draw from the stock and must try to create a new sequence that either matches the original lead suit or creates a new lead suit for the next player to follow.

### Strategic Trump Swap (New)
Players aim to follow the suit of the lead card, but can exchange a card from their hand for a trump card from a common pool if they cannot follow suit. The new trump becomes the suit to follow in subsequent turns if chosen.

### Sequential Suit Shift (New)
Players must follow the suit of the lead card in a cascading manner, where each turn shifts the suit requirement clockwise among the players' hands. If unable to comply, they must contribute to a central pot, which is won by the first player to successfully follow suit again.

### Suit Cycle Bluff (New)
Players strive to follow the suit of the lead card but can declare an alternate suit, bluffing to force others into drawing from the stock. If challenged and caught bluffing, they must draw additional cards, but if successful, the bluffed suit becomes the new lead for the next round.

### Adaptive Suit Sequence (New)
Players follow the suit of the lead card until someone cannot and draws from the stock, initiating a temporary 'joker' round. During this round, any suit can be played but does not change the lead, adding strategic depth to maintain or regain control.

### Revolving Suit Mastery (New)
Players follow the suit of the lead card with a twist: if they exhaust all cards of a particular suit, they gain a bonus action to swap it with a card from the stock, changing the active lead suit for the next player, introducing layers of strategy in conserving suits.
\end{lstlisting}
\end{tcolorbox}

\subsection{User Study on Game Variation}
\label{sup:mechanics-user-study}

For each method listed in Table 1 (Vanilla, PromptBreeder, and our approach), we selected 6 game variants generated from the base game ``Go Fish'', resulting in a total of 18 variants. To reduce the reading burden for participants, we summarized each variant’s differences from the base game. Three examples are shown below:

\begin{tcolorbox}[
breakable,
title=Go Fish variants from our work (Differences summarized),  
colframe=darkgray, 
colback=white,
coltitle=white,
left=1mm, 
right=1mm, 
top=0mm, 
bottom=0mm, 
fonttitle=\small,
boxrule=0.2mm, 
]
\begin{lstlisting}[language=markdown]
**Variation 1**
The varied game of Go Fish introduces a sequential accumulation advantage, allowing players to peek at the top card of the stock pile if they successfully request and complete sequential ranks within the same turn. Unlike the base game, the varied game focuses more on strategic card requests and offers players the opportunity to gain insight into upcoming cards. The varied game emphasizes book completion as players can continue their turn by requesting sequential ranks, adding a new layer of strategy. Additionally, while most rules regarding initial dealing and turn sequences remain consistent, the varied game subtly shifts the focus of strategy towards leveraging sequential requests and peeking privileges.

**Variation 2**
In the Varied Game: Position Swap Edition of Go Fish, a notable difference is the introduction of a Position Swap mechanic, allowing players to strategically swap cards by correctly guessing two card ranks in an opponent's hand. Unlike the base game, this version emphasizes strategic plays and interactions over straightforward card requests. While both versions involve forming books by collecting four cards of the same rank, the varied game allows for an additional guess-based action that can change the course of a player's strategy. The varied game also permits shared victories in the event of a tie, whereas the base game typically ends in a draw or a singular winner.

**Variation 3**
In the varied version of Go Fish, the stockpile cards are entirely hidden, unlike in the base game where the number of cards remaining is known. The varied game introduces a "misdirection" special ability, allowing a player once per cycle to request a card rank they do not have, at the risk of skipping their next turn if caught misusing it more than once. A cycle in the varied game is defined as a complete sequence where each player has taken a turn, resetting the misdirection ability. Finally, winning conditions are similar, but misdirection use and penalties during the game add an extra layer of strategy not present in the base game.
\end{lstlisting}
\end{tcolorbox}

In the user study, 12 participants were first provided with a full description of the base game. They were then presented with 9 variant summaries in text (3 from each method), randomly sampled from the total pool of 18. Participants rated each summary based on the following criteria, using a 5-point scale:
\begin{itemize}
    \item Logical correctness: if the variant contains logically correct game mechanics.
    \item Strategic engagement: Inspired by Agon (competition) in Caillois's forms of play, which refers to the engagement derived from strategic rivalry.
    \item Luck engagement: Inspired from Caillois's Alea (chance), which refers to the engagement elicited by elements of chance or randomness.
    \item Overall creativity, rated as follows: (1) Not creative at all; (2) Lacking sufficient creativity; (3) Some creative ideas worth refining; (4) Mostly creative but requires polish; (5) Highly creative.
\end{itemize}

We used a linear mixed-effects model to evaluate the effects of different methods. Results in Table~\ref{tab:user-rating} indicate that all three approaches achieved satisfactory levels of logical correctness (scores above 3.5). The table uses our method as the reference, which means the intercept is the average score for our method, and the coefficients for PromptBreeder and Vanilla represent the difference in scores compared to our method. Our method matched PromptBreeder in creativity (both scoring around 3.5), and outperformed the vanilla LLM baseline by 0.5 points. Given that our approach also demonstrated higher novelty (as shown in Table 1), we argue that it can produce game variants that are both objectively novel and subjectively creative.

\begin{table*}[htbp]
\centering
\caption{\textbf{User ratings on generated game variants across methods}. Mixed-effects model estimates for the effect differences of other baselines against our method.}
\label{tab:user-rating}
\begin{tabular}{lrrrrrr}
\hline
Term & Coef. & Std.Err. & z & P$>|z|$ & [0.025 & 0.975] \\
\hline
\multicolumn{7}{l}{\textbf{Logical Correctness}} \\
Intercept              & 4.056 & 0.115 & 35.165 & 0.000 & 3.830 & 4.282 \\
Effect of PromptBreeder& 0.000 & 0.200 & 0.000 & 1.000 & -0.392 & 0.392 \\
Effect of Vanilla      & -0.250& 0.200 & -1.252 & 0.211 & -0.642 & 0.142 \\
Scale                  & 0.359 &       &        &       &       &       \\
Log-Likelihood         & -136.9899 &   &        &       &       &       \\
\hline
\multicolumn{7}{l}{\textbf{Strategic Engagement}} \\
Intercept              & 3.583 & 0.207 & 17.300 & 0.000 & 3.177 & 3.989 \\
Effect of PromptBreeder& -0.722& 0.267 & -2.700 & 0.007 & -1.246& -0.198 \\
Effect of Vanilla      & -0.639& 0.257 & -2.489 & 0.013 & -1.142& -0.136 \\
Scale                  & 0.776 &       &        &       &       &       \\
Log-Likelihood         & -177.4349 &   &        &       &       &       \\
\hline
\multicolumn{7}{l}{\textbf{Luck Engagement}} \\
Intercept              & 2.639 & 0.193 & 13.655 & 0.000 & 2.260 & 3.018 \\
Effect of PromptBreeder& 0.528 & 0.219 & 2.412 & 0.016 & 0.099 & 0.957 \\
Effect of Vanilla      & -0.056& 0.237 & -0.234 & 0.815 & -0.521& 0.410 \\
Scale                  & 0.676 &       &        &       &       &       \\
Log-Likelihood         & -170.2324 &   &        &       &       &       \\
\hline
\multicolumn{7}{l}{\textbf{Creativity}} \\
Intercept              & 3.528 & 0.125 & 28.180 & 0.000 & 3.282 & 3.773 \\
Effect of PromptBreeder& -0.167& 0.053 & -3.143 & 0.002 & -0.271& -0.063 \\
Effect of Vanilla      & -0.444& 0.196 & -2.264 & 0.024 & -0.829& -0.060 \\
Scale                  & 0.610 &       &        &       &       &       \\
Log-Likelihood         & -164.7739 &   &        &       &       &       \\
\hline
\end{tabular}
\end{table*}

Furthermore, our method tends to generate strategically engaging games better than the other two approaches by a margin of 0.6. While, PromptBreeder shows relative strength in producing luck-driven games, leading by 0.5.

\subsection{More Generated Varied Games}
\label{sup:game-variants}

Our generated variant, \emph{Hearts of Time}, differs from classical \emph{Hearts} mainly by introducing the Temporal Suit Shift, a once-per-game ability that lets a player declare a temporary trump suit for a trick, adding a new layer of strategic timing.

\begin{tcolorbox}[
breakable,
title=Variant of Heart from our work,  
colframe=darkgray, 
colback=white,
coltitle=white,
left=1mm, 
right=1mm, 
top=0mm, 
bottom=0mm, 
fonttitle=\small,
boxrule=0.2mm, 
]
\begin{lstlisting}[language=markdown]
# Hearts of Time Game System Ruleset (Refined)

### 1. **Game State**

#### **Common Information:**
- **Trick History:** Sequence of played cards in each trick, visible to all players.
- **Current Leader:** Player who won the last trick, visible to all players.
- **Turn Information:** Current player and actions taken during the turn.
- **Player Order:** Logical order of play, visible to all players.

#### **Player-Specific Information:**
- **Public**:
  - **Scores:** Current interference (disturbance scores) known at all times.
  - **Trick Wins:** Number of tricks each player has won.
- **Private**:
  - **Player Hand:** Cards held by the player.
  - **Temporal Suit Shift Status:** Whether the player has used their shift ability.

---

### 2. **Card**

#### **Attributes:**
- **Rank:** From 2 to Ace.
- **Suit:** One of {Hearts, Diamonds, Clubs, Spades}.
- **Special Cards:**
  - **Queen of Spades:** 13 points of disturbance.
  - **Hearts:** Each incurs 1 point of disturbance.

---

### 3. **Deck and Initial Dealing**

#### **Deck Composition:**
- A standard 52-card deck divided into four suits.

#### **Initial Dealing:**
- **4 Players:** Each player receives 13 cards.
- Other player counts require deck adjustment by removing low-ranking cards, maintaining suit balance.

---

### 4. **Legal Action Space**

#### **On a Turn, a Player May:**
1. **Lead a Trick:**
   - Play any card from their hand.
   - Pre-requisite: Must follow suit, unless void in that suit.
2. **Follow a Trick:**
   - Play a card matching the led suit, if possible.
   - Else, play any card.
3. **Declare Temporal Suit Shift:**
   - Announce the shift of rule for a trick to establish a temporary trump suit. This can be declared before playing any card during a player's turn.
   - Pre-requisite: Can only be used once per game per player.

---

### 5. **Round**

#### **Sequence of Play:**
1. The player with the 2 of Clubs starts the first trick.
2. Players take turns in clockwise order, leading with the suit following rules.
3. Each trick consists of:
   - Playing cards in the order of player turns.
   - Using Temporal Suit Shift if strategically advantageous. When declared, the player's chosen suit acts as trump for that entire trick.
4. The winner of a trick leads the next trick.
5. Play continues until all cards have been played.

#### **Winning Conditions:**
- Game ends when a player reaches a preset disturbance threshold (commonly 100 points).
- Player with the lowest disturbance score wins.

---

### 6. **Other Game Mechanics & Rules**

- **No Hearts or Queen of Spades First Trick:** During the first trick, players must avoid playing hearts or the Queen of Spades unless they have no alternative.
- **Shooting the Moon:** Capturing all Hearts and Queen of Spades reduces disturbance to zero for that round, increasing opponents' scores by 26 points.
- **Temporal Strategy:** Deciding when to use Temporal Suit Shift is critical to altering trick outcomes. The temporary trump suit impacts only one trick and reverts after its resolution.

---

### 7. **Player Observation Information**

#### **Visible Information to Each Player:**
- Cards in their own hand.
- Cards played in tricks.
- Scores of each player.
- Temporal Suit Shift use and declared trump suit.

#### **Hidden Information:**
- Cards in opponents' hands.

---

### 8. **Payoffs**

##### **Endgame Scoring:**
- A point of disturbance equals 1 for each Heart.
- Queen of Spades is worth 13 disturbance points.
- Shooting the Moon results in zero points for the round, with opponents' scores increased.

##### **Winning Player's Reward:**
- Having the lowest disturbance score leads to victory when the game concludes.

With these refinements, the "Hearts of Time" game provides clarity on the use and impact of Temporal Suit Shift, maintaining the intrigue and strategy native to Hearts, while clearly outlining the new rules. Enjoy your journey through the mystical realm as you guide your temporal path among the "Hearts of Time."
\end{lstlisting}
\end{tcolorbox}

Our variant of classical \emph{Hollywood Eights} builds on the original rules of matching ranks, suits, or playing wild eights. However, when the classical version revolves around matching suits or ranks and using wild eights to control flow until a player empties their hand, the variant introduces a sequence merging mechanics and a scoring race to 150 points.

\begin{tcolorbox}[
breakable,
title=Variant of Hollywood Eights from our work,  
colframe=darkgray, 
colback=white,
coltitle=white,
left=1mm, 
right=1mm, 
top=0mm, 
bottom=0mm, 
fonttitle=\small,
boxrule=0.2mm, 
]
\begin{lstlisting}[language=markdown]
## Card Merge Challenge: Refined Ruleset

### 1. **Game State**

#### **Common Information:**
- **Stock Pile:** Cards remaining in the draw pile.
- **Starter Pile:** Top card visible to all players.
- **Player Order:** Determined at the beginning of the game, visible to all.

#### **Player-Specific Information:**
- **Public:**
  - **Merged Sequences:** Sequences successfully completed by all players and visible to all.
  - **Scoreboard:** Cumulative points of each player.
- **Private:**
  - **Player Hand:** Cards held by each player, visible only to themselves.

---

### 2. **Card**

#### **Attributes:**
- **Rank:** One of {A, 2, 3, 4, 5, 6, 7, 9, 10, J, Q, K}.
- **Suit:** One of {Hearts, Diamonds, Clubs, Spades}.
- **Unique Symbol:** A special symbol on each card for merging purposes.
- **Special Card:** Eight acts as a wild card for both suit and symbol changes.

---

### 3. **Deck and Initial Dealing**

#### **Deck Composition:**
- A standard 52-card deck with additional unique symbols on each card.

#### **Initial Dealing:**
- Deal five cards face down to each player.
- Remaining cards form the stock pile.
- Top card of the stock pile is turned up to start the starter pile. If this is an eight, it is inserted into the deck, and another card is drawn.

---

### 4. **Legal Action Space**

#### **On a Turn, a Player May:**
1. **Play a Card:**
   - Place one card from their hand on the starter pile.
   - Prerequisite: The card must match the top card by rank, suit, or symbol.
2. **Merge Sequence:**
   - Create a sequence with the top card of the starter pile combined with cards from the player's hand.
   - Required: Sequence of three or more cards sharing a common symbol.
   - Additional Rule: Only one merged sequence can be created per turn.
3. **Use Wild Card (Eight):**
   - Play an eight to change suit and symbol trajectory.
4. **Draw a Card:**
   - Draw from the stock if they can't play, OR opt to strategically draw despite alternatives.
   - Prerequisite: No matching card in hand, or by strategic choice.

---

### 5. **Round**

#### **Sequence of Play:**
1. Player left of the dealer starts.
2. Turns proceed in clockwise order.
3. During a turn, a player:
   - Plays a card or merges.
   - Draws a card (if needed).
   - The turn ends and passes to the next player.
4. Play continues until:
   - A player reaches 150 points through merges and regular gameplay.

#### **Winning Conditions:**
- A player wins upon reaching 150 points through successful merges and card play.

---

### 6. **Other Game Mechanics & Rules**

- **Merge Bonus:** Successfully merging a sequence grants a 10-point bonus per card in the sequence.
- **Wild Card Usage:** Eights alter game flow, allowing suit and symbol changes for new sequence formation.
- **Score Recording:** Results recorded across up to four ongoing games, scoring detailed sequentially.

---

### 7. **Player Observation Information**

#### **Visible Information to Each Player:**
- The top card of the starter pile.
- Merged sequences on the table.
- Cumulative scores of all players.

#### **Hidden Information:**
- Identity of cards in the stock pile.
- Cards in opponents' hands.

---

### 8. **Payoffs**

#### **Endgame Scoring:**
- Points awarded for each merged sequence:
  - Each successful merge contributes to the score based on the number of cards in the sequence.
  - Align scoring with the traditional logic: Face cards = 10 points; Eights = 20 points; Ace = 15 points; Numbered cards = Pip value.

#### **Objective Achievement:**
- Game winner is the first to score 150 points through strategic merging and regular gameplay, leveraging visible and hidden game states smartly.

This refined description ensures the gameplay mechanics are well-aligned with the new merging objective and special card function, providing clarity and ensuring strategic depth.
\end{lstlisting}
\end{tcolorbox}

\section{Code Generation}
\label{sup:code}

\begin{figure*}[ht]
  \centering
  \includegraphics[width=\linewidth]{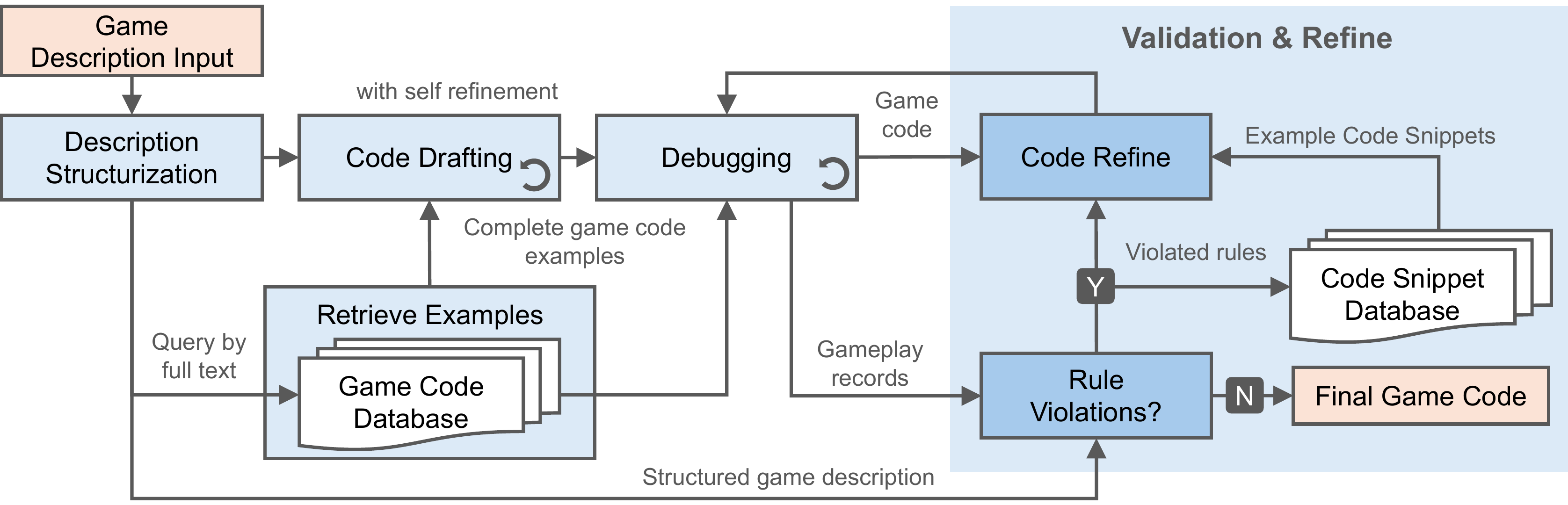}
  \caption{LLM agent workflow of game code generation.}\label{fig:agentflow}
\end{figure*}

As shown in Figure~\ref{fig:agentflow}, the pipeline comprises two processes: initial code generation and iterative validation, both relying on interaction with our custom card game engine. 

This section starts from introducing the system design of the game engine in~\ref{sup:game-engine}, followed by details of the first process in~\ref{sup:init-code}. Validation process has been briefly described in Section~\ref{sec:game-code}. In~\ref{sup:code-validate}, we provide generated intermediate results for qualitative assessment. Finally, we evaluate the effectiveness of our metric ``PCon'' in \ref{sup:pcon-eval}.

\subsection{Game engine design}
\label{sup:game-engine}

Instead of generating entire game code from scratch as in \citet{chatdev}, we instruct LLMs to fill in several predefined functions, which are integrated into our card game engine in Python.

\begin{figure}[ht]
  \centering
  \includegraphics[width=.48\textwidth]{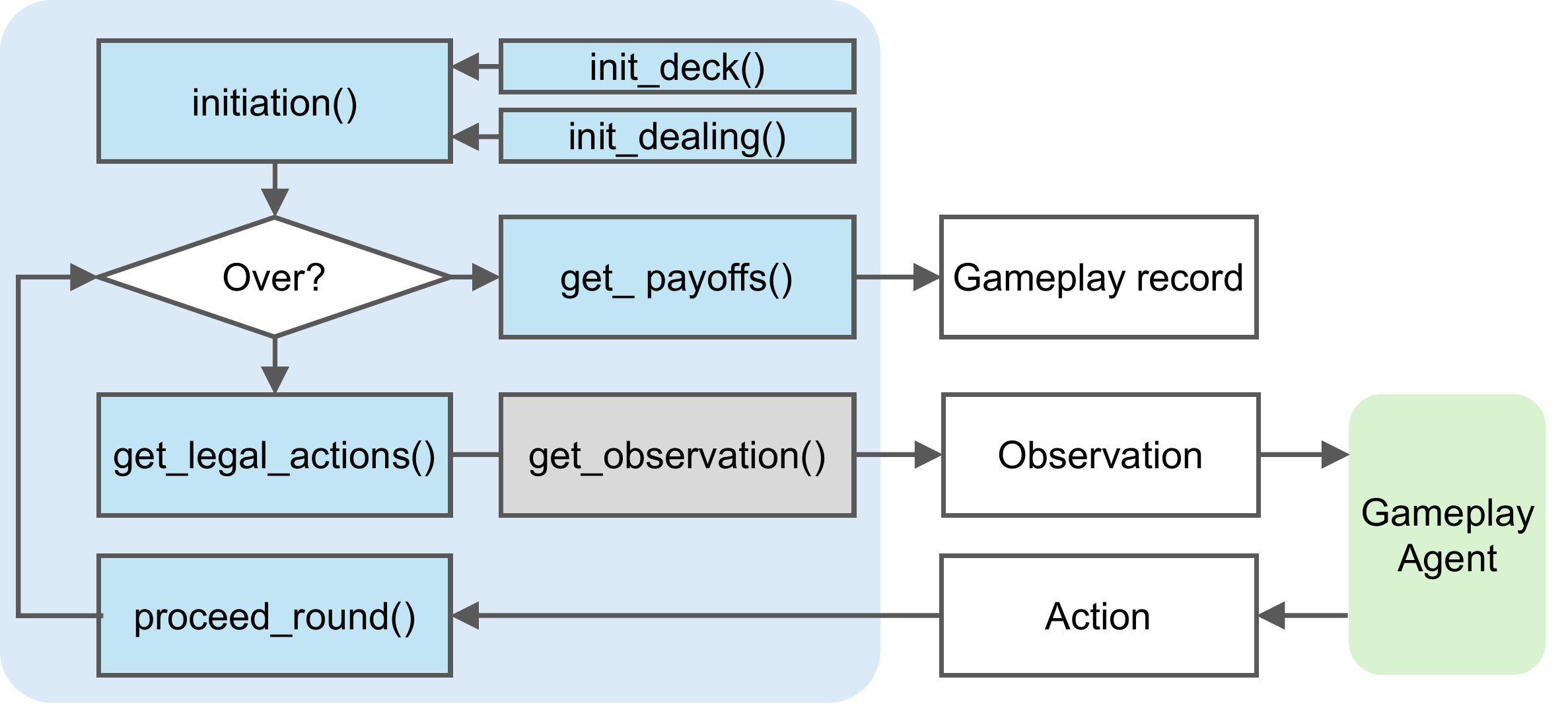}
  \caption{\textbf{Game Engine Framework Design}. The game engine, composed of 6 core functions (in light blue) in a fixed flow, interacts with external gameplay agent (in light green) through game observation and game action in dictionary forms.}\label{fig:game-engine}
\end{figure}

To enable maximum compatibility for various card games and easy integration with gameplay AI, we adopt a framework structure where (1) game state updates and gameplay AI are decoupled \cite{zha_douzero_2021}; and (2) dependencies among game logics are minimized by a functional programming design \cite{wu2024instructiondriven}.
The logic of card games are abstracted into 6 functions (initiation, initialize deck, initial dealing, proceed round, get legal actions, and get payoffs) running in a predefined procedure (see Figure \ref{fig:game-engine}). Each function, which modifies the game state in dictionary form, is what our pipeline will generate.

The gameplay AI is integrated into this system by receiving game observations and outputting game actions in dictionaries. While our gameplay AI directly handles dictionaries, for all LLM-based agents we compared against, the game state is converted to the following format. The example below shows the observation information in one game turn. LLM agents may receive concatenated information from the past several turns.

\subsection{Initial Code Generation}
\label{sup:init-code}

Given the text description of a card game as input, this process creates its game code without syntax or runtime error by the following phases:

\paragraph{Description structurization} LLM is called to rewrite the description into a structured markdown format, where game description is sectioned into game states, deck initialization, legal action space, game play process, etc. 

\paragraph{Retrieve examples} By comparing the text embedding cosine similarity between the structured description of the input and the structured game descriptions in the example database. We can retrieve the closest game examples with their corresponding code.

In our work, we construct an example database containing $5$ game descriptions and their game code (``crazy-eights", ``california-jack", ``boat-house-rum", ``bull-poker", ``baccarat"). These games are selected from the $106$ games without overlaps to our test set.

\paragraph{Code Drafting} The framework drafts game code using the structured description, game engine code with function templates (detailed in \ref{sup:game-engine}), and retrieved code examples. Those supporting information is included in the system prompt as an in-context learning practice. After the code draft undergoes self-refinements with a fixed number of rounds, the code is sent to the next phase.

\paragraph{Repeated Debugging} The game code is iteratively refined using feedback from the game engine to eliminate runtime errors. During testing, player choices are randomized by sampling the legal action space generated by the `get legal actions' component. The game is executed 10 times with different random seeds to explore diverse game states. If runtime errors occur, the error messages are fed back to the LLM, which adjusts the code accordingly. This process repeats until the game runs without errors.

\subsection{Intermediate Results in Validation}
\label{sup:code-validate}

In the code validation process, our method compares the generated gameplay record with the game description, identifying whether there exists an inconsistency. Multiple consecutive observations from different players are concatenated together.

\begin{tcolorbox}[
breakable,
title=Gameplay Record Section for ``Go Fish'',  
colframe=darkgray, 
colback=white,
coltitle=white,
left=1mm, right=1mm, top=0mm, bottom=0mm,
fonttitle=\small,
boxrule=0.2mm, 
]
\begin{lstlisting}[language=markdown]
Player 0 observes: 
```
# Common Information
num_players: 4
current_player: 0
books: [3, 1, 3, 3]
stock_size: 0

# Player Information

## Player 0 (Self)
current_player: True
hand: [K, 10, 10, K, K]
books: ['2', '9', '5']

## Player 1
hand_size: 1
books: ['4']

## Player 2
hand_size: 2
books: ['6', '3', '7']

## Player 3
hand_size: 4
books: ['Q', 'J', 'A']

# Legal Actions
0: request-(target_player: 1, rank: K)
1: request-(target_player: 1, rank: 10)
2: request-(target_player: 2, rank: K)
3: request-(target_player: 2, rank: 10)
4: request-(target_player: 3, rank: K)
5: request-(target_player: 3, rank: 10)
```

Player 0 decides to: request-(target_player: 1, rank: 10)
----------
Player 1 observes: 
```
# Common Information
num_players: 4
current_player: 1
books: [3, 1, 3, 3]
stock_size: 0

# Player Information

## Player 0
hand_size: 5
books: ['2', '9', '5']
...
\end{lstlisting}
\end{tcolorbox}

Relfecting on the gameplay record and the game description, LLM generates validation analysis as below:

\begin{tcolorbox}[
breakable,
title=Consistency validation results,  
colframe=darkgray, 
colback=white,
coltitle=white,
left=1mm, right=1mm, top=0mm, bottom=0mm,
fonttitle=\small,
boxrule=0.2mm, 
]
\begin{lstlisting}[language=markdown]
***Step by step evaluation***

1. **Turn 0 (Player 0):**
   - Player 0 has cards [K, 10, 10, K, K] and chooses to request from Player 1 the rank 10. This is legal because Player 0 has cards of rank 10.
   - Player 1 has 1 card; no information if it was 10. Assuming it wasn't, Player 0 should "Go Fish," but the stock is empty. The turn ends.

2. **Turn 1 (Player 1):**
   - Player 1 has only card 8 and requests rank 8 from Player 0.
   - Since Player 0 does not have this card (as per observed hand [K, 10, 10, K, K]), the request fails, but again no stock to draw from. The turn ends.

3. **Turn 2 (Player 2):**
   - Player 2 has cards [10, 10] and requests rank 10 from Player 0. This action is valid since they have rank 10 cards.
   - Player 0 has two 10s to pass over-indicating a successful request. Player 2 should take another turn but this is not visible.

4. **Turn 3 (Player 3) and Subsequent Observations:**
   - Player 3 requests rank 8 from Player 1. Since Player 1 does not have the rank (hand size is 1, card 8 not explicitly confirmed), the turn would typically result in "Go Fish."
   - The same cycle continues; player's actions become clearer, but crucial observations like drawn or obtained cards are missing, suggesting errors in turn execution.

5. **Game End:**
   - Legal actions remaining for the player with hand [K, K, K] show remaining possibilities accurately. No correct handling of potential book formation or reattempts captured.

***Analysis Summary***
Summary:
```text
The progression of the turns does not effectively simulate correct sequence dictated by rules, notably in request fulfillment and subsequent actions not following successful card claims due to incomplete turn handling. Data missing for card exchanges leads to incongruity with established gameplay mechanics.
```
Quote (optional):
```markdown
"Each turn consists of: Requesting cards from an opponent; Drawing from the stock if the request results in 'Go Fish'; Optionally revealing a book if completed; Passing the turn after drawing or failing to make a catch. Re-attempt: If a player successfully gets cards, they can continue asking for the same or different rank."
```
\end{lstlisting}
\end{tcolorbox}

Based on the validation results, code edits are proposed by LLM. Same as \cite{agentless}, partial replacement edits are returned to save LLM token usage.

\begin{tcolorbox}[
breakable,
title=Code Edit Proposal Example,  
colframe=darkgray, 
colback=white,
coltitle=white,
left=1mm, right=1mm, top=0mm, bottom=0mm,
fonttitle=\small,
boxrule=0.2mm, 
]
\begin{lstlisting}[]
<<<<<<< SEARCH
def proceed_round(action: dict, game_state: Dict, logger) -> DotDict:
    """Process the action and update the game state."""
    current_player = game_state.common.current_player
    logger.info(f"Player {current_player} takes an action: {action}")
    
    if action['action'] == 'request':
        process_request(action, game_state, logger)
    elif action['action'] == 'draw':
        if game_state.common.facedown_cards.stock:
            card = game_state.common.facedown_cards.stock.pop()
            game_state.players[current_player].facedown_cards.hand.append(card)
            logger.info(f"Player {current_player} draws a card.")
        else:
            logger.info("Stock pile is empty, no card to draw.")

    check_for_books(game_state.players[current_player], game_state, logger)
    
    # Check if the game is over
    if is_game_over(game_state, logger):
        return game_state
    
    game_state.common.current_player = next_player(game_state, logger)
    return game_state
=======
def proceed_round(action: dict, game_state: Dict, logger) -> DotDict:
    """Process the action and update the game state."""
    current_player = game_state.common.current_player
    logger.info(f"Player {current_player} takes an action: {action}")
    
    if action['action'] == 'request':
        if process_request(action, game_state, logger):
            return game_state  # Successful request allows player to continue.
    elif action['action'] == 'draw':
        if game_state.common.facedown_cards.stock:
            card = game_state.common.facedown_cards.stock.pop()
            game_state.players[current_player].facedown_cards.hand.append(card)
            logger.info(f"Player {current_player} draws a card.")
        else:
            logger.info("Stock pile is empty, no card to draw.")

    check_for_books(game_state.players[current_player], game_state, logger)
    
    # Check if the game is over
    if is_game_over(game_state, logger):
        return game_state
    
    game_state.common.current_player = next_player(game_state, logger)
    return game_state
>>>>>>> REPLACE
\end{lstlisting}
\end{tcolorbox}

\subsection{Effectiveness of PCon Metric}
\label{sup:pcon-eval}

To verify whether LLMs are good at consistency evaluation, We assess our method, PCon, through three experiments:
\begin{itemize}
    \item Ground Truth Evaluation: We used 5 manually verified game codes, each with 5 gameplay records. PCon assessed their consistency with the original game descriptions, where we expected near-perfect scores.
    \item Small Deviations: Gameplay records were compared to game descriptions (chosen from our dataset) with 95\% cosine similarity to the original. We expected high but slightly reduced consistency scores.
    \item Large Deviations: Records were compared to game descriptions with 70\% similarity, where we expected significantly lower scores.
\end{itemize}

\begin{figure}[htbp]
  \centering
  \includegraphics[width=.48\textwidth]{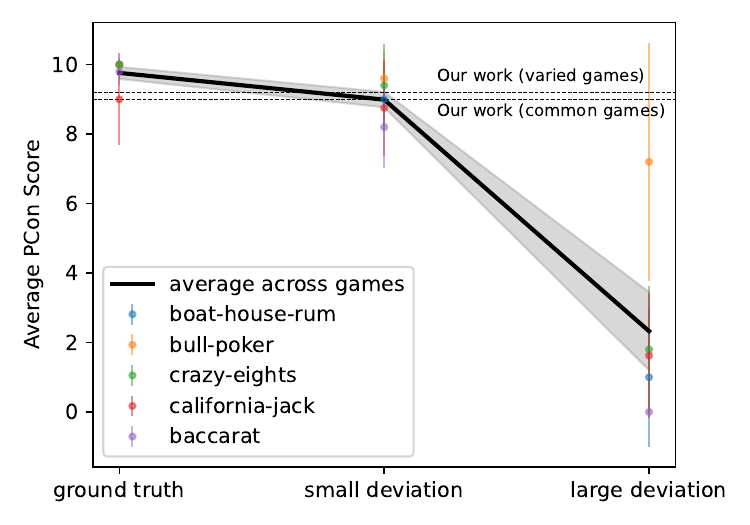}
  \caption{\textbf{Effectiveness of PCon metric}. Average PCon metrics in three game sets, showing PCon is aligned with the actual consistency between the game play record and game description. Gray shaded areas represent the standard errors of average PCon.}\label{fig:pcon-eval}
\end{figure}

The results aligned with our expectations (see Figure~\ref{fig:pcon-eval}). The average PCon score for the ground truth group ($9.68 \pm 0.85$) was significantly higher than both deviation groups ($8.96 \pm 1.24$ in small deviation group and $2.25 \pm 3.17$ in large deviation group, $p<0.05$ in paired t-test), indicating PCon effectively captures consistency, even in challenging scenarios involving subtle deviations.

It is important to note that LLM-based validation process improves the code alignment, rather than ensuring 100\% correct code. We conducted a manual verification to all generated games where LLM detects no misalignment, where we still find minor inconsistencies. For instance, there are simplified implementation of complex rules: for poker games with a large number of card evaluation standards (such as Straight Flush, Four of a kind, Full House, etc.), there are cases where our method does not implement all the standards mentioned in the game description.
Also, incorrect information visibilities may occur: sometimes, private information for specific players may be leaked to the public by game state outputs.
However, the validation process still minimizes human effort in code development, which benefits the overall prototyping process.

\section{Gameplay AI}
\label{sup:gameai}

\begin{figure}[htbp]
    
  \centering
    \includegraphics[width=.75\textwidth]{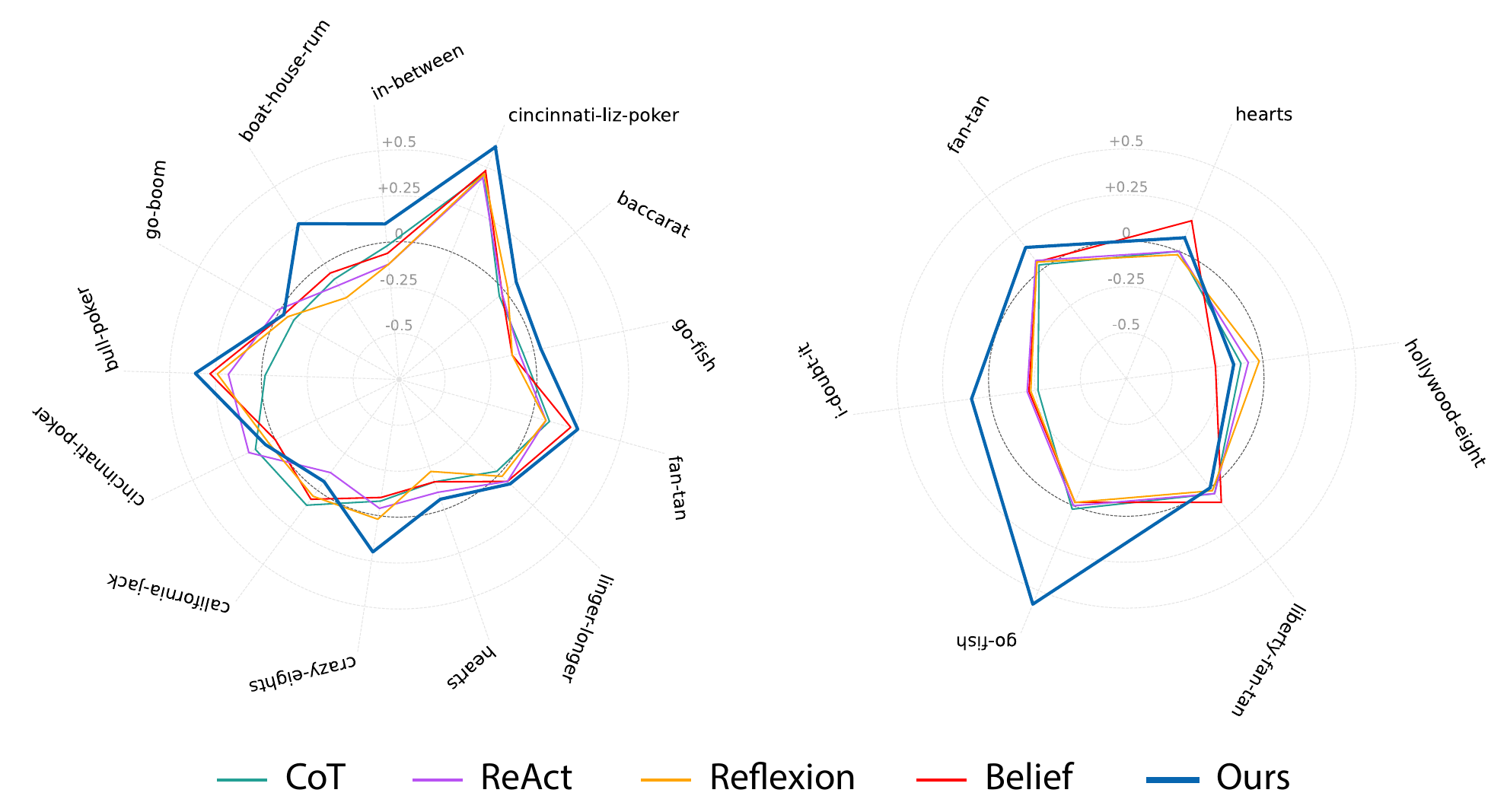}

  \caption{\textbf{Gameplay AI win rate advantage by games}. Win rate advantage to random agents when different methods are applied. The performance of random agents ($A(p,p_{random})=0$) is shown as black dotted circles. (left) Common games (right) varied games, where the names of base games are shown in radial axis. }\label{fig:win-adv-random}
  \label{fig:adv_win}

\end{figure}

Figure \ref{fig:adv_win} provides a detailed breakdown of the win rate advantage across individual games. The plot uses a radial layout, with each axis corresponding to a specific game and advantage scores plotted as distances from the center. Our method spans the largest area in both the common and varied game sets, indicating broad performance gains. In the common games, it achieves the highest win rate advantage in 10 out of 13 games, demonstrating strong coverage across diverse rule sets.

All games used in our evaluation are listed in Table~\ref{tab:games}. 

\begin{table}[htbp]
    \centering
    \begin{tabular}{@{}p{0.23\linewidth}p{0.7\linewidth}@{}}
        \hline
        Genre & Games \\
        \hline
        Rummy & Boat House Rum \\
        Casino & Baccarat, Bull Poker, Cincinnati Poker, 
                 Cincinnati Liz Poker, Fan Tan, Fan Tan (Varied), 
                 In-Between, Liberty Fan Tan (Varied) \\
        Trick Taking & California Jack, Crazy Eights, Go Boom, 
                       Hearts, Hearts (Varied), Hollywood Eights (Varied) \\
        Other & Go Fish, Go Fish (Varied), Linger Longer, 
                I Doubt it (Varied) \\
        \hline
    \end{tabular}
    \caption{Card games used in the evaluation, grouped by genre}
    \label{tab:games}
\end{table}

\subsection{Compare to Human-designed Heuristics}
\label{sup:gameai-heuristics}

We compared our method to two heuristic agents from \citet{zha_rlcard_2020}, using head-to-head win rate of the last game play agent as the main metric. Across different player orderings, our approach matched or exceeded human-designed heuristics: outperforming them in Leduc Hold'em, underperforming in Uno, but still surpassing the random agent.

\begin{table}[htbp]
\centering
\begin{tabular}{llcc}
\hline

Proceeding& Last & Leduc Holdem & Uno \\
\hline
Random&Random & 70.5 & 24.5 \\
Heuristic&Random & 79.3 & 18.5 \\
Ours&Random & 39.7 & 23.4 \\
Random&Heuristic & 70.7 & 27.8$^{*}$ \\
Random&Heuristic & 78.5 & 24.4$^{*}$ \\
Ours&Heuristic & 48.1$^{*}$ & 29.8$^{*}$ \\
Random&Ours & 82.0$^{*}$ & 25.8 \\
Heuristic&Ours & 84.6$^{*}$ & 21.4$^{*}$ \\
Heuristic&Ours & 59.3$^{*}$ & 24.1 \\
\hline
\end{tabular}
\caption{Win Rate between human heuristics and our work in Leduc Holdem and Uno. $^{*}$ indicates a significant difference from settings where random player is the last one. (chi-square test, $p < 0.05$).}
\end{table}

These results show our method can generate competent game AI suitable for early-stage prototyping and generalizes to game variants without manual design.

\subsection{Ablation Study}
\label{sup:gameai-ablation}

To assess the contribution of key components in our method, we perform an ablation study comparing the full agent against two simplified variants: 
\begin{itemize}
\item No Ensemble (``$\times$ens''): It generates a single comprehensive heuristic function instead of ensembling multiple LLM-generated strategies.
\item No Optimization (``$\times$opt''): It retains the full set of heuristics generated before the augmentation step but skips the optimization phase, using all heuristics equally without selection. 
\end{itemize}

As shown in Table~\ref{tab:gameai-adv-ablation}, both ablations result in a sharp drop in performance, neither achieves a positive average advantage over random agents. These results indicate that both ensembling and the optimization-based policy refinement are essential to the effectiveness of our gameplay AI. All scores are averaged over 50 runs per game.

\begin{table}[htbp]
    \centering
    \sisetup{separate-uncertainty= true, table-format=3.1(1), multi-part-units=single}
    \begin{tabular}{p{0.6cm}SSS}
        \hline
               & \text{Common} & \text{Mutated} & \text{All Games} \\
        \hline
        Ours   & \color{blue} 14.7 \pm 19.3 & \color{blue}  19.7 \pm 30.5 & \color{blue} 16.3 \pm 22.6 \\
        $\times$ens  & -3.2 \pm 7.0& -2.1 \pm 9.2~$\dagger$ &  -2.9 \pm 7.5~$\dagger$ \\
        $\times$opt  & 0.2 \pm 9.9 & -1.5 \pm 6.2~$\dagger$ & -0.4 \pm 8.7~$\dagger$ \\
        \hline
    \end{tabular}
    \caption{\textbf{Game AI win rate advantage between our method and its ablations}. Means and standard deviations (shown after $\pm$) of win rate advantages ($\times 10^{-2}$) across all game instances. All metrics between our method and its ablations show statistically significant difference (p-value $\leq 0.05$ in paired t-test). $\dagger$ represents two metrics are unlikely to be different (p-value $\geq 0.7$ in paired t-test).}
    \label{tab:gameai-adv-ablation}
\end{table}

\subsection{Sensitivity Analysis}
\label{sup:sensitivity-policy-num}

We vary the number of base policy items to evaluate its impact on gameplay AI performance, measured by win rate advantage. 
As shown in Table~\ref{tab:sensitivity-n}, using a small number of base policy items (e.g., $n=2$) results in decreased performance, likely due to reduced diversity in the policy pool, where critical strategies may not be surfaced due to sampling limitations or combinatorial constraints. However, increasing the number yields diminishing returns. We observe no significant performance difference between our default setting ($n=4$) and a larger configuration ($n=6$). This plateau may reflect the limitations of our current method: the LLM may fail to generate meaningfully diverse strategies beyond a certain point, or improved policies may not emerge from simple combinations of heuristic functions derived from those strategies. 

\begin{table}[htbp]
    \centering
    \begin{tabular}{lccc}
        \hline
         Population      & \text{Common} & \text{Varied} & \text{All games} \\
        \hline
        n=2 & 8.5 \tiny{$\pm 15.5$} & 10.0 \tiny{$\pm 14.4$} & 8.9 \tiny{$\pm 14.8$} \\
        n=4 (Ours) &  \color{blue} 14.6 \tiny{$\pm 19.3$} & \color{blue}  19.7 \tiny{$\pm 30.5 $} &\color{blue} 16.3 \tiny{$\pm 22.6$}\\
        n=6 & 16.5 \tiny{$\pm 17.5$} & 17.3 \tiny{$\pm 21.8 $} & 16.7 \tiny{$\pm 18.3$} \\
        \hline
    \end{tabular}
    \caption{\textbf{Game AI win rate advantage with different base policy population}. Means and standard deviations (shown after $\pm$) of win rate advantages ($\times 10^{-2}$) over random agents across all game instances. 
    }
    \label{tab:sensitivity-n}
\end{table}

\subsection{Baseline Implementation}
\label{sup:game-ai-eval}

Although some previous work demonstrates satisfactory performance in several popular games, they are excluded from our comparative analysis as their primary contributions focus on specific game genres \cite{wang2023avalon} or particular attributes of game mechanics \cite{light2024strategist}. 
For example, methods that learn directly from reward signals at MCTS leaf nodes \cite{light2024strategist} can hardly work on games with extremely unbalanced search trees. Because it is challenging to reach all terminal states with reasonable computational resources. As a result, one has to estimate the reward without reaching the terminal nodes, which is beyond the scope of their work. 

Consequently, we focus on prior work that are scalable. We choose Chain-of-Thought~\cite{CoT}, ReAct~\cite{yao_react_2023}, Reflexion~\cite{shinn_reflexion_2023}, and AgentPro~\cite{zhang2024agentpro} as our baselines. As AgentPro is particularly noteworthy for its application of Belief-aware Decision-Making imperfect information game scenarios, we specifically implement the Belief-aware part for comparison.

Although for our evaluation, we intended to use the most performant LLM model available for every method being compared, the number of LLM calls being made by other works is a polynomial order of magnitude greater than our work. For the number of games being compared, this token cost is prohibitively expensive (Figure~\ref{fig:token_cost}). We have therefore opted for the most performant LLM within budget for other works (4o-mini, which boasts similar performance) and 4o for our work. Our model is limited in that it requires at least 4o for its coding accuracy. However, despite the difference between LLM backbones, the 4o-mini being used for other works is comparable if not better than the LLM backbones that were originally used in the respective works.

\subsubsection{Inputs for Baseline Methods}

While our work uses game state dictionary as inputs, all baseline methods require natural language inputs. Therefore, we adopt the gameplay record format from the code generation pipeline to describe the self game state, with private information about other players removed.

To improve readability and highlight the public progression of the game, we also include automatically generated game comments alongside the self game state. Rather than being generated directly by LLMs, game comments are generated by the commentary functions inside the game code. 
The commentary function example below, designed by LLMs, fetch variables from game state dictionary to compose game comments such as ``Player 2 creates a fusion card from: 2-diamond, 3-diamond, 4-diamond''.

\begin{tcolorbox}[
breakable,
title=Commentary Function Example,  
colframe=darkgray, 
colback=white,
coltitle=white,
left=1mm, right=1mm, top=0mm, bottom=0mm,
fonttitle=\small,
boxrule=0.2mm, 
]
\begin{lstlisting}[]
if can_create_fusion(cards, player):
    for card in cards:
        player.private.player_hand.remove(card)
    player.public.fusion_cards.append(cards)
    logger.info(f"Player {current_player} creates a fusion card from: {cards}.")
    if check_fusion_victory(player):
        game_state.common.is_over = True
        game_state.common.winner = current_player
        logger.info(f"Player {current_player} wins by creating three fusion cards!")
\end{lstlisting}
\end{tcolorbox}

Game comments are accumulated over the course of an entire turn, recording the public actions and effects of other players. Together, the comments and the self game state form a composite game observation for a single turn. For each agent, we maintain a game history consisting of the five most recent turns.

\begin{tcolorbox}[
breakable,
title=Game Observation for Boat House Rum,  
colframe=darkgray, 
colback=white,
coltitle=white,
left=1mm, 
right=1mm, 
top=0mm, 
bottom=0mm, 
fonttitle=\small,
boxrule=0.2mm, 
]
\begin{lstlisting}[]
# Gameplay since your last decision
Dealing 7 cards to each player
Player 0 decides to: draw-(source: discard_pile)
Player 0 draws cards from discard pile.
Player 0 decides to: discard-(card_idx: 6)
Player 0 discards A-diamonds.
Player 1 decides to: draw-(source: discard_pile)
Player 1 draws cards from discard pile.
Player 1 decides to: discard-(card_idx: 3)
Player 1 discards 9-diamonds.

# Common Information
num_players: 3
current_player: 2
has_drawn_cards_this_turn: False
discard_pile: [{'rank': '9', 'suit': 'diamonds'}]
stock_size: 30

# Player Information

## Player 0
score: 0
melds: []
hand_size: 7
recent_discard_draw_size: 0

## Player 1
score: 0
melds: []
hand_size: 7
recent_discard_draw_size: 0

## Player 2 (Self)
score: 0
current_player: True
melds: []
hand: [{'rank': '3', 'suit': 'diamonds'}, {'rank': 'Q', 'suit': 'spades'}, {'rank': 'Q', 'suit': 'hearts'}, {'rank': '3', 'suit': 'spades'}, {'rank': '4', 'suit': 'hearts'}, {'rank': '4', 'suit': 'spades'}, {'rank': '2', 'suit': 'diamonds'}]
recent_discard_draw: []

# Legal Actions
0: draw-(source: stock)
1: draw-(source: discard_pile)
\end{lstlisting}
\end{tcolorbox}

\subsubsection{Training Reflexion Agents}

The evaluation protocol involved training Reflexion agents across all games, with performance measured using a rolling average winning rate calculated over windows of 40 games. This measurement was repeated for 10 distinct windows across all games in our test set. The optimal reflection step was subsequently determined by selecting the iteration that yielded the highest mean winning rate across the entire game suite.

\subsection{Intermediate Results}
\label{sup:gameai-int-results}

During the construction of gameplay AI, our method produces policy components in text form first, which is converted to code during the second augmentation process.
\begin{tcolorbox}[
breakable,
title=Gameplay AI policy component as texts,  
colframe=darkgray, 
colback=white,
coltitle=white,
left=1mm, 
right=1mm, 
top=0mm, 
bottom=0mm, 
fonttitle=\small,
boxrule=0.2mm, 
]
\begin{lstlisting}[language=markdown]
**Prioritize Building Melds**
Focus on arranging cards in the hand into matched sets of either three or four of a kind, or sequences of three or more cards of the same suit. Discard cards that do not contribute towards completing a meld.

**Draw from Discard Pile Judiciously**
Only draw from the discard pile when the top two cards can be directly used to form a meld or significantly improve the hand. Otherwise, prefer drawing from the stock pile to maintain unpredictability.

**Reserve Key Cards**
Keep hold of high-value cards like aces and face cards only if they are part of an ongoing strategy to complete a meld or if they are necessary to win the game soon. Otherwise, discard them to minimize penalty points.

**PotentialMeldCreationValue**
A metric that assesses the potential of forming new melds after drawing two cards, either from the stock pile or discard pile. It evaluates the likely improvement in the hand's structure, considering both the draw options and the current meld opportunities.

**UnmatchedCardPenalty**
A metric that evaluates the penalty incurred from holding high-value unmatched cards. This metric helps in deciding the priority of discarding high-value cards from the hand to lower potential score penalties.

**Refined Meld Building Strategy** (Mutually inspired policy)
To refine the strategy for building melds, the AI player should incorporate the following steps: prioritize drawing from the source that maximizes 'PotentialMeldCreationValue.' When choosing cards to discard, evaluate based on 'UnmatchedCardPenalty,' preferably discarding high-penalty cards unless they significantly contribute to potential melds. Use 'MeldCompletionLikelihood' to assess laying off opportunities. Lastly, factor in 'OpponentInterferenceRisk' to prevent assisting opponents with your discards. These criteria ensure strategic decision-making based on the game state, maximizing winning potential and minimizing penalties.
\end{lstlisting}
\end{tcolorbox}

\begin{tcolorbox}[
breakable,
title=Gameplay AI policy component in code,  
colframe=darkgray, 
colback=white,
coltitle=white,
left=1mm, 
right=1mm, 
top=0mm, 
bottom=0mm, 
fonttitle=\small,
boxrule=0.2mm, 
]
\begin{lstlisting}[]
def score(state: dict, action: str) -> float:
    # Helper function to calculate the potential meld creation value
    def calculate_potential_meld_creation_value(hand, card):
        value = 0
        # Check for potential sets (three or four of a kind)
        rank_count = sum(1 for c in hand if c['rank'] == card['rank'])
        if rank_count == 2: # possibility to make a set of three
            value += 0.3
        if rank_count == 3: # possibility to make a set of four
            value += 0.5
                    
        # Check for potential sequences
        card_ranks = {'A': 1, '2': 2, '3': 3, '4': 4, '5': 5, '6': 6,
                      '7': 7, '8': 8, '9': 9, '10': 10, 'J': 11, 'Q': 12, 'K': 13}
        hand_ranks = [card_ranks[c['rank']] for c in hand if c['suit'] == card['suit']]
        hand_ranks.append(card_ranks[card['rank']])
        hand_ranks.sort()
        
        # Look for consecutive cards
        for i in range(len(hand_ranks) - 2):
            if hand_ranks[i+1] == hand_ranks[i] + 1 and hand_ranks[i+2] == hand_ranks[i] + 2:
                value += 0.4
                
        return value
        
    # Helper function to calculate unmatched card penalty
    def calculate_unmatched_card_penalty(hand):
        penalty = 0
        for card in hand:
            penalty += {'A': 11, 'K': 10, 'Q': 10, 'J': 10,
                        '10': 10, '9': 9, '8': 8, '7': 7,
                        '6': 6, '5': 5, '4': 4, '3': 3, '2': 2}[card['rank']]
        return penalty

    # Extract relevant information
    current_player_index = state['common']['current_player']
    current_player = state['players'][current_player_index]
    hand = current_player['facedown_cards']['hand']
    discard_pile = state['common']['faceup_cards']['discard_pile']

    # Initialize the result score
    result_score = 0.0

    # Calculate potential meld creation value of the action
    if action == "draw":
        draw_source = state['legal_actions'][0]['args']['source']  # assuming we're provided with actions arguments
        
        # Check both stock and discard pile for card drawing
        if draw_source == "stock":
            # Assuming uncertainty based draw, hence less score boost
            result_score += 0.1  # a relatively small optimistic boost for drawing
        elif draw_source == "discard_pile":
            if discard_pile:
                top_discard_card = discard_pile[-1]
                result_score += calculate_potential_meld_creation_value(hand, top_discard_card)
            else:
                result_score += 0.05  # minor score addition when discard pile is empty

    # Evaluating penalty for unmatched cards
    unmatched_penalty = calculate_unmatched_card_penalty(hand)
    result_score -= unmatched_penalty * 0.01  # reduce score proportionally to the penalty

    # Make sure that the final score is between 0 and 1\n    result_score = max(0, min(1, result_score))

    return result_score
\end{lstlisting}
\end{tcolorbox}

\section{Benefits for related studies}

Our work benefits the related research communities in the following ways: 

\paragraph{Game distance and recommendation}

Compared to game distance metrics that builds on low-level code structure \cite{distancemapping2021}, our work could facilitate a more explainable distance metric, as its node units incorporates higher-level abstractions with increased legibility.
Also, as our graph-indexed database reveals the dependency between game mechanics (in Figure \ref{fig:mechanic_dependencies}), it can be integrated into a game mechanics recommendation framework with friendly user interface as \cite{pitako}, thus enabling a human-in-the-loop game design.

\paragraph{Game Code Data Generator} Our work can be treated as a synthetic data generator specialized in creating programs with long and complex instruction. Its generated game code can be used to fine-tune domain specific code generative models \cite{wu2024instructiondriven}, or be used as test case data for general program synthesis methods.

\paragraph{Extendable Gameplay AI Benchmark}
Compared to prior work that focus on a small set of card games \cite{costarelli2024gamebench,zha_douzero_2021}, we provide a much larger scale game benchmark environment for gameplay AIs. 
Additionally, prior work in LLM-based gameplay AI suffers a limitation in evaluation: LLMs may have seen the game strategy during their training, making it challenge to evaluate the true performance of the gameplay framework. By designing and constructing novel game environments with minimal human effort, our work can help bridging this gap.



\section{LLM System Prompts}
\subsection{Game Mechanics Design}

\begin{tcolorbox}[
breakable,
title=System Prompt for game logic extraction (depth 1),  
colframe=promptcolor, 
colback=white,
left=1mm, right=1mm, top=0mm, bottom=0mm, fonttitle=\small, boxrule=0.2mm, 
]
\begin{lstlisting}[]
You are a wonderful card game designer who extract game logic chains from the game description. You will be given a game description and you need to extract the game logic chains step by step.

# Task

Read the following game description and answer the questions: how to win this card game? Please respond with all direct mechanics that contribute or hinder to this. for example: discard all the cards, get highest hand score. Your output should be a JSON object with the following format:
```json
[
    {
        "name": "Name of the game mechanic",
        "type": "<only choose among: Contribute, Hinder, Mixed>",
        "description": "A concise explanation of how the mechanic works",
        "reasoning": "Explain how this mechanic contributes to the goal"
    },
    ...
]
```

Remember to extract mechanics that DIRECTLY relate to the goal. Examples:
```
input: <a UNO game description>
correct extraction: "Empty the hands"
wrong extraction: "Play a matching card" or "Draw 4 cards".
analysis: "Play a matching card" or "Draw 4 cards" should be extracted in future steps.
```
```
input: The main purpose of the game is to remove all cards from the table, assembling them in the tableau before removing them. Initially, 54 cards are dealt to the tableau in ten piles, face down except for the top cards. The tableau piles build down by rank, and in-suit sequences can be moved together. The 50 remaining cards can be dealt to the tableau ten at a time when none of the piles are empty.
correct extraction: "Remove all cards from the table"
wrong extraction: "Move sequences strategically".
```

# Game Description
{game_description}
\end{lstlisting}
\end{tcolorbox}

\begin{tcolorbox}[
breakable,
title=System Prompt for game logic extraction (depth 2+),  
colframe=promptcolor, 
colback=white,
left=1mm, right=1mm, top=0mm, bottom=0mm, fonttitle=\small, boxrule=0.2mm, 
]
\begin{lstlisting}[]
What are the game mechanics/concept that directly trigger/stop this mechanic: {goal}? Please respond with the direct mechanic that contributes to this. 

- Try your best to extract ALL direct mechanics.
- When evaluating, pay special attention to mechanics that are interesting or unique.
- You can choose from previously extracted mechanics: {previous_mechanics}. But you can also extract new mechanics/concepts.
- Use the same format as before. 
- Usually there are always mechanics or concepts to extract, unless you have reached the end of the chain such as card design or very basic elements.
- If there are no more mechanics or concepts to extract, respond with:
```json
[]
```
\end{lstlisting}
\end{tcolorbox}

\begin{tcolorbox}[
breakable,
title=System Prompt for reflecting on logic extraction,  
colframe=promptcolor, 
colback=white,
left=1mm, right=1mm, top=0mm, bottom=0mm, fonttitle=\small, boxrule=0.2mm, 
]
\begin{lstlisting}[]
Check and refine your results based on the following criteria:
(1) Are there any mechanics that are not DIRECTLY related to the goal?

Respond a complete result in the same format as before.
\end{lstlisting}
\end{tcolorbox}

\begin{tcolorbox}[
breakable,
title=System Prompt for concept summary,  
colframe=promptcolor, 
colback=white,
left=1mm, right=1mm, top=0mm, bottom=0mm, fonttitle=\small, boxrule=0.2mm, 
]
\begin{lstlisting}[]
**Role**: You are an assistant tasked with summarizing game concept from a list of gameplay mechanics.

**Instructions**:
1. **Input**: You will receive a list of gameplay concept instances. These concepts are often centered around a theme or have similar mechanics.
2. **Output**: Your response should include:
```json
{
    "name": "<The name of the game concept>",
    "description":{
        "common": "<A concise statement that generalizes the core concept or theme common across the listed gameplay concept instances.>",
        "variation": "How the description of given instances vary from the shared common themes."
    }
}
```

# Your input
{input_text}
\end{lstlisting}
\end{tcolorbox}

\begin{tcolorbox}[
breakable,
title=System Prompt for designing new logic instances,  
colframe=promptcolor, 
colback=white,
left=1mm, right=1mm, top=0mm, bottom=0mm, fonttitle=\small, boxrule=0.2mm, 
]
\begin{lstlisting}[]
{prefix_prompt}. Now design the novel concept instance based on your summary. 
- The concept should center around the summary you have concluded.
- The concept should be different from all given instances.
- You should specify its name and description. 
- You should follow this format:
```json
{
    "name": "Name of the game concept (e.g., Dynamic Wild Card, Critical Threshold).",
    "description": "A short concise explanation, following the writing style of the previous instances."
}
```
\end{lstlisting}
\end{tcolorbox}

\begin{tcolorbox}[
breakable,
title=Prefix Prompts for designing new logic instances (partially shown),  
colframe=promptcolor, 
colback=white,
left=1mm, right=1mm, top=0mm, bottom=0mm, fonttitle=\small, boxrule=0.2mm, 
]
\begin{lstlisting}[]
"You are tasked with designing a novel card game concept. Reflect on all aspects of this challenge, including understanding core elements, researching existing games, defining game objectives, developing themes, innovating mechanics, player interaction, rule-setting, prototyping, gathering feedback, and iterating on your design. Confirm your comprehension before proceeding with design suggestions.",

"Before diving into the problem, take a moment to clear your mind and approach it with a fresh perspective. With a clear mind, you can create more innovative solutions.\n\nYou are a skilled card game designer. With these fresh insights, please design a novel game concept instance based on your experience and expertise.",

"As an experienced designer of card games, create an innovative game concept inspired by your prior insights.",

"To design a novel card game concept, first ensure you have a clear understanding of the essential elements: game objectives, rules, mechanics, and player dynamics. Identify what information might be missing or unclear, such as game balance issues, potential for player engagement, or themes. Use this review to innovate and develop an improved or entirely new game concept that enhances the player's experience and meets the outlined objectives. Focus on creating unique mechanics or themes that differentiate it from existing games.",

"You are a skilled card game designer. Please design a novel game concept instance, emphasizing teamwork and open communication. Seek input on the implications and potential impact of different design elements on gameplay. Leverage the diverse perspectives and expertise of the group to enhance the creativity and effectiveness of the game design.",   
\end{lstlisting}
\end{tcolorbox}

\subsection{Code Generation}
\begin{tcolorbox}[
breakable,
title=Prompts for description structurization,  
colframe=promptcolor, 
colback=white,
left=1mm, right=1mm, top=0mm, bottom=0mm, fonttitle=\small, boxrule=0.2mm, 
]
\begin{lstlisting}[]
Design a structured ruleset for implementing a card game system based on the provided input. Ensure the output includes key components as below. The output should be comprehensive, logical, and organized in a format suitable for programming or detailed documentation purposes. Wrap the output in a markdown block.

Include the following sections:
1. **Game State**
   - Define the game state, categorized into common information and player-specific information (grouped into public and private).
2. **Card**
    - Specify card attributes such as rank, suit, and any special abilities or values.
3. **Deck and Initial Dealing**
    - Describe the deck composition, dealing process, and setup at the beginning of the game.
4. **Legal Action Space**
    - List all possible actions players can perform during their turn, specifying the prerequisites of each action.    
5. **Round**
    - Describe the sequence of play and how the game progresses from one player to the next.
    - Elaborate in each players' turn, the order of actions they can take, and the outcomes of each action.
    - Explain how the game ends and the winning conditions. Pay attention to corner cases such as deck exhaustion or all players passing.
6. **Other Game Mechanics & Rules**
    - Detail any additional game mechanics, rules, or special actions that players can take during the game.
7. **Player Observation Information**
    - Specify what information players can observe during the game, such as their hand, the starter pile, declared suits, and opponent actions.
8. **Payoffs**
    - Explain when game ends, how scoring works, including point values for cards.

Ensure clarity and precision to facilitate implementation or usage as a reference for game rules.

# Example
{example}
\end{lstlisting}
\end{tcolorbox}

\begin{tcolorbox}[
breakable,
title=System Prompts for code draft,  
colframe=promptcolor, 
colback=white,
left=1mm, right=1mm, top=0mm, bottom=0mm, fonttitle=\small, boxrule=0.2mm, 
]
\begin{lstlisting}[]
You are a card game programmer tasked with implementing a card game based on the given description. Using the provided class templates, your goal is to write only the necessary child classes and implement only the methods indicated with `TODO` comments.

**Instructions:**
- Include only the methods you need to override from the provided class templates.
- Respond with complete, runnable Python code.
- Do **not** include TODOs, placeholders, or explanations; output only the final code.

**Code Environment:**
- These code belongs to a larger game framework. Use them as reference only. Don't include them in your response.
```
{environment_code}
```

**Code Template:**
- Only modify or implement the methods specified with `TODO` comments to complete the game logic.
```
{python_classes}
```

**Examples for Reference:**  
Use these examples as a guide for response format and method implementation.
{examples}

---

### Your Task
Based on the following game description, implement the required classes and methods:

**Game Description:**
```
{game_description}
```

### Note:
- Do **not** raise exceptions (e.g., `ValueError`) when parsing action strings. Instead, ensure the legal action space and action string format are appropriately structured.
\end{lstlisting}
\end{tcolorbox}

\begin{tcolorbox}[
breakable,
title=Prompts for consistency validation,  
colframe=promptcolor, 
colback=white,
left=1mm, right=1mm, top=0mm, bottom=0mm, fonttitle=\small, boxrule=0.2mm, 
]
\begin{lstlisting}[]
You are a card game programmer who verifies code for a card game. You are given a card game description and a part of game play log using the code.

# Task
- You should evaluate step by step to see if the game play log aligns with the rules in the game description.
- Also, examine if the legal action choices in each turn is correct and complete.
- If the game play aligns with the rules, simply return "pass" in the analysis summary.
- If the game play does not align with the rules, you should response in a two-part format: summary and quote(optional). Focus one issue at a time.

# Your game description
```
{game_description}
```

# Your game play log
Note: Only the last several turns of the play log is provided. But if the play log is too short or empty, there might be some errors in the game code.
```
{game_play_log}
```

# Output Format

If the game play log aligns with the rules:
```
***Step by step evaluation***
<your evaluation here>

***Analysis Summary***
```pass```
```

If you doubt the log is too short or empty because of some errors in the game code:
```
***Step by step evaluation***
<your evaluation here>

***Analysis Summary***
```log is too short or empty```
```

Otherwise:
```
***Step by step evaluation***
<your evaluation here>

***Analysis Summary***
Summary:
```text
<summarize the issue>
```
Quote (optional):
```markdown
<quote related game description segment if game play log does not align with the rules>
```
```
\end{lstlisting}
\end{tcolorbox}

\subsection{Gameplay AI Generation}
\begin{tcolorbox}[
breakable,
title=Design game policy components (strategy style) in text,  
colframe=promptcolor, 
colback=white,
left=1mm, right=1mm, top=0mm, bottom=0mm, fonttitle=\small, boxrule=0.2mm, 
]
\begin{lstlisting}[]
You are a powerful assistant who designs an AI player for a card game.

# Game rules
{game_description}

# Game state
The AI player knows all cards in its hands, all game play history. But it does not know the content of other players' hands.

# Potential actions
{game_actions}

# Task
Please think in steps to provide me {item_num} useful strategies to win the game.
For each strategy, please describe its definition and how it relates to the game state and a potential action.

# Response format
Please respond in the following JSON format:
{format_instructions}
\end{lstlisting}
\end{tcolorbox}

\begin{tcolorbox}[
breakable,
title=Design game policy components (metric style) in text,  
colframe=promptcolor, 
colback=white,
left=1mm, right=1mm, top=0mm, bottom=0mm, fonttitle=\small, boxrule=0.2mm, 
]
\begin{lstlisting}[]
You are a powerful assistant who designs an AI player for a card game.

# Game rules
{game_description}

# Game state
The AI player knows all cards in its hands, all game play history. But it does not know the content of other players' hands.

# Potential actions
{game_actions}

# Task
To design a good game play policy, we need to design some game state metrics that constitute a reward function.
Now please think in steps to tell me what useful metric can we derive from a game state?
The metric should be correlated with both the game state and the potential action. Provide me with {item_num} metrics.

# Response format
Please respond in the following JSON format:
{format_instructions}
\end{lstlisting}
\end{tcolorbox}

\begin{tcolorbox}[
breakable,
title=Mutually inspire game policy components in text,  
colframe=promptcolor, 
colback=white,
left=1mm, right=1mm, top=0mm, bottom=0mm, fonttitle=\small, boxrule=0.2mm, 
]
\begin{lstlisting}[]
You are a powerful assistant who designs an AI player for a card game.

# Game rules
{game_description}

# Game state
The AI player knows all cards in its hands, all game play history. But it does not know the content of other players' hands.

# Potential actions
{game_actions}

# Task
Given the following strategy of the game:
```json
{game_strategy}
```

Please think in steps to refine the strategy using the following criteria:
(1) If the strategy has anything obscure, for example, if it mentions "strategically use" or "use at critical moments" without specifying what the critical moments are, please clarify what the critical moments are.
(2) If the strategy is conditioned on a game state metric, please describe how such a strategy will be conditioned on the game state. Here are some hints of the game state:
```json
{game_metrics}
```

# Response format
Please respond in the following JSON format:
{format_instructions}
\end{lstlisting}
\end{tcolorbox}

\begin{tcolorbox}[
breakable,
title=Design code for game policy components,  
colframe=promptcolor, 
colback=white,
left=1mm, right=1mm, top=0mm, bottom=0mm, fonttitle=\small, boxrule=0.2mm, 
]
\begin{lstlisting}[]
You are an action-value engineer trying to write action-value functions in python. Your goal is to write an action-value function that will help the agent decide actions in a card game.

# The game
{game_description}

# The policy
In this action-value function, you will focus on the following policy of the game:
{game_policy}

# The input
The function should be able to take a game state and a planned game action as input. The input should be as follows:
{input_description}

# The output
You should return a reward value ranging from 0 to 1. It is an estimate of the probability of winning the game. 
The closer the reward is to 1, the larger chance of winning we will have.
Try to make the output more continuous.
The reward should be calculated based on both the game state and the given game action.

# Response format
You should return a python function in this format:
```python
def score(state: dict, action: str) -> float:
    pass
    return result_score
```
\end{lstlisting}
\end{tcolorbox}

\begin{tcolorbox}[
breakable,
title=Refine code for game policy components,  
colframe=promptcolor, 
colback=white,
left=1mm, right=1mm, top=0mm, bottom=0mm, fonttitle=\small, boxrule=0.2mm, 
]
\begin{lstlisting}[]
Here are some criteria for the code review:
- No TODOs, pass, placeholders, or any incomplete code;
- Include all code in the score function. Don't create custom class or functions outside;
- the last line should be "return result_score", and the result_score should be a float;
- You can only use the following modules: math, numpy (as np), random;
- no potential bugs;

First, you should check the above criteria one by one and review the code in detail. Show your thinking process.
Then, if the codes are perfect, please end your response with the following sentence:
```
Result is good.
```

Otherwise, you should end your response with the full corrected function code. 
\end{lstlisting}
\end{tcolorbox}

\begin{tcolorbox}[
breakable,
title=Game-play Prompt for Chain-of-thought Agent,  
colframe=promptcolor, 
colback=white,
left=1mm, right=1mm, top=0mm, bottom=0mm, fonttitle=\small, boxrule=0.2mm, 
]
\begin{lstlisting}[]
You are a player in a card game. Please do your best to beat the other players and win the game.

The card game is as follows:
```
{game_description}
```

Your current state is as follows:
{observation}

Please think in steps and make a decision based on the current state of the game.

You should return an index of the action you want to take from the list of legal actions.
Wrap your response in a dictionary with the key 'action' and the value as the index of the action you want to take.
for example (You MUST return the action index in the following format):

```json
{
    "action": 0
}
```
\end{lstlisting}
\end{tcolorbox}

\begin{tcolorbox}[
breakable,
title=Game-play Prompt for ReAct Agent,  
colframe=promptcolor, 
colback=white,
left=1mm, right=1mm, top=0mm, bottom=0mm, fonttitle=\small, boxrule=0.2mm, 
]
\begin{lstlisting}[]
You are a player in a card game. Please do your best to beat the other players and win the game.

The card game is as follows:
```
{game_description}
```

Previous observation of game history and your current state is as follows:
```
{observation}
```

Your available actions are as follows:
```
{actions}
```

Please think in steps and make a decision based on the current state of the game.
You should return an index of the action you want to take from the list of legal actions.

Instructions:
1. Analyze the query, previous reasoning steps, and observations.
2. Decide on the next action: choose an action or provide a final answer.
3. Respond in the following JSON format:

Remember:
- Be thorough in your reasoning.
- Choose actions when you need more information.
- Always base your reasoning on the actual observations from chosen action.

If you have enough information to answer the query, wrap your response in a dictionary with the key 'action' and the value as the index of the action you want to take.
for example (You MUST return the action index in the following format):

```json
{
    "thought": "Your detailed reasoning about what to do next",
    "action": 0
}
```
\end{lstlisting}
\end{tcolorbox}

\begin{tcolorbox}[
breakable,
title=Game-play Prompt for Reflexion Agent,  
colframe=promptcolor, 
colback=white,
left=1mm, right=1mm, top=0mm, bottom=0mm, fonttitle=\small, boxrule=0.2mm, 
]
\begin{lstlisting}[]
You are a player in a card game. Please do your best to beat the other players and win the game.

The card game is as follows:
```
{game_description}
```

Previous observation of game history and your current state is as follows:
```
{observation}
```

Your available legal actions are as follows:
```
{actions}
```

Your previous reflection based on the game state is as follows:
```
{reflection}
```

Please think in steps and make a decision based on the current state of the game.
You should return an index of the action you want to take from the list of legal actions.

Instructions:
1. Analyze the query, previous reasoning steps, and observations.
2. Decide on the next action: choose an action or provide a final answer.
3. Respond in the following JSON format:

Remember:
- Be thorough in your reasoning.
- Choose actions when you need more information.
- Always base your reasoning on the actual observations from chosen action.

If you have enough information to answer the query, wrap your response in a dictionary with the key 'action' and the value as the index of the action you want to take.
You MUST return the action index in the following format:

```json
{
    "thought": "Your detailed reasoning about what to do next",
    "action": 0
}
```
\end{lstlisting}
\end{tcolorbox}

\begin{tcolorbox}[
breakable,
title=Long-term Reflection Prompt for Reflexion Agent,  
colframe=promptcolor, 
colback=white,
left=1mm, right=1mm, top=0mm, bottom=0mm, fonttitle=\small, boxrule=0.2mm, 
]
\begin{lstlisting}[]
You are a player in a card game. The card game is as follows:
```
{game_description}
```
This is the payoffs for the game:
```
[{payoffs}], you are player {idx} (list index starting from 0).
```
Here is the previous reflections based on the game results:
```
{reflection}
```

The game is over. Please reflect on the game and provide a detailed summarization for WINNING the game including strategies used, reasoning for actions, and any other relevant information.
You should return a string with your reflection on the game, including strategies used, reasoning for actions, and any other relevant information.
Each reflection item should be precised and concise, NEVER repeat the same reflection twice.
You MUST return the reflection in the following format and ONLY add at most 2 most important new summarization to the previous reflection:
```json
{
    "reflection": "1: strategy 1; 2: strategy 2; ..."
}
```
\end{lstlisting}
\end{tcolorbox}

\begin{tcolorbox}[
breakable,
title=Game-play Prompt for Belief Agent,  
colframe=promptcolor, 
colback=white,
left=1mm, right=1mm, top=0mm, bottom=0mm, fonttitle=\small, boxrule=0.2mm, 
]
\begin{lstlisting}[]
You are a player in a card game. Please do your best to beat the other players and win the game.

The card game description is as follows:
```
{game_description}
```

Previous observation of game history and your current state is as follows:
```
{observation}
```

Here is your previous reflection based on the game state is as follows:
```
{reflection}
```

Previous belief is as follows:
```
self-belief: {self_belief}
world-belief: {world_belief}
```


Please read the game decription and observation carefully.
Then you should analyze your own cards and your strategies in Self-belief and then analyze the opponents cards in World-belief.
Lastly, please select your action from:
```
{actions}
```

You MUST return the action index in the following format:
```json
{
    "self-belief": "I have a good hand", 
    "world-belief": "Dealer has a bad hand", 
    "action": 1
}
```
\end{lstlisting}
\end{tcolorbox}

\stopcontents[mytoc] 

\end{document}